\documentclass[12pt,a4paper]{article}
\pdfoutput=1
\usepackage[utf8]{inputenc}
\usepackage{natbib}
\usepackage[english]{babel}
\usepackage{color} 
\usepackage{xcolor,colortbl}
\usepackage[center]{caption}
\usepackage{amsmath}
\usepackage{bbm}
\usepackage{afterpage}
\usepackage{dsfont}
\usepackage{amsfonts}
\usepackage{amssymb}
\usepackage{graphicx}
\usepackage{subfig}
\usepackage{enumitem}
\usepackage{changepage}
\usepackage[colorlinks=true,citecolor=black,linkcolor=black,urlcolor=black]{hyperref}
\usepackage{authblk}
\usepackage{float}
\usepackage{fullpage}
\usepackage[justification=centering]{caption}
\usepackage{lineno}
\usepackage{eurosym}
\usepackage[margin=10pt,font=small,labelfont=bf]{caption}
\usepackage{multirow,booktabs,setspace,caption}
\usepackage{scalerel,stackengine}
\usepackage{rotating}
\usepackage{bm}

\usepackage[utf8]{inputenc} 
\usepackage[T1]{fontenc}    
\usepackage{url}            
\usepackage{booktabs}       
\usepackage{amsfonts}       
\usepackage{nicefrac}       
\usepackage{microtype}      

\usepackage{float}          
\restylefloat{table}        
\usepackage{makecell}
\usepackage{multirow}
\usepackage{subfig}

\usepackage{graphicx}

\usepackage{epstopdf}
\newcommand{\f}[1]{\textbf{#1}}
\usepackage{rotating}
\usepackage{amsmath}

\makeatletter
\newcommand\footnoteref[1]{\protected@xdef\@thefnmark{\ref{#1}}\@footnotemark}
\makeatother

\DeclareCaptionLabelSeparator*{spaced}{\\[2ex]}
\newcommand{\bdelta}{\bm{\delta}}
\newcommand{\x}{\bm{x}}
\newcommand{\z}{\bm{z}}
\newcommand{\g}[1]{\textcolor{gray}{#1}}
\DeclareMathOperator*{\E}{\mathbb{E}}
\DeclareMathOperator*{\median}{\text{median}}
\DeclareMathOperator*{\CE}{\text{CE}}
\author[a,1,2]{Evgenia~Rusak*}
\author[a,1,2]{Lukas~Schott*}
\author[a,1]{Roland~S.~Zimmermann*}
\author[b,1]{Julian Bitterwolf}
\author[b,1]{Oliver~Bringmann\textsuperscript{$\dagger$}}
\author[a,1]{Matthias~Bethge\textsuperscript{$\dagger$}}
\author[a,1]{Wieland~Brendel\textsuperscript{$\dagger$}}

\affil[*]{equal contribution}
\affil[$\dagger$]{joint senior authors}
\affil[a]{first.last@bethgelab.org}
\affil[b]{first.last@uni-tuebingen.de}
\affil[1]{University of T\"ubingen}
\affil[2]{International Max Planck Research School for Intelligent Systems}

\font\myfont=cmr12 at 20.0pt

\begin{document}
\title{\myfont A simple way to make neural networks robust against diverse image corruptions}

\date{}

\maketitle
\begin{abstract}
   The human visual system is remarkably robust against a wide range of naturally occurring variations and corruptions like rain or snow. 
   In contrast, the performance of modern image recognition models strongly degrades when evaluated on previously unseen corruptions. 
   Here, we demonstrate that a simple but properly tuned training with additive Gaussian and Speckle noise generalizes surprisingly well to unseen corruptions, easily reaching the previous state of the art on the corruption benchmark ImageNet-C (with ResNet50) and on MNIST-C. We build on top of these strong baseline results and show that an adversarial training of the recognition model against uncorrelated worst-case noise distributions leads to an additional increase in performance. 
   This regularization can be combined with previously proposed defense methods for further improvement. 
\end{abstract}

\section{Introduction}
    While Deep Neural Networks (DNNs) have surpassed the functional performance of humans in a range of complex cognitive tasks \citep{he2016deep, DBLP:journals/corr/XiongDHSSSYZ16a, Silver2017MasteringTG, Campbell:2002:DB:512148.512152, OpenAI_dota}, they still lag behind humans in numerous other aspects. One fundamental shortcoming of machines is their lack of robustness against input perturbations. Even minimal perturbations that are hardly noticeable for humans can derail the predictions of high-performance neural networks.
    
    For the purpose of this paper, we distinguish between two types of input perturbations. One type are minimal image-dependent perturbations specifically designed to fool a neural network with the smallest possible change to the input. These so-called \textit{adversarial perturbations} have been the subject of hundreds of papers in the past five years, see e.g.\ \citep{szegedy2013intriguing, madry2018towards, schott2018towards, DBLP:journals/corr/abs-1801-02774}. Another, much less studied type are \textit{common corruptions}. These perturbations occur naturally in many applications and include simple Gaussian or Salt and Pepper noise; natural variations like rain, snow or fog; and compression artifacts such as those caused by JPEG encoding. All of these corruptions do not change the semantic content of the input, and thus, machine learning models should not change their decision-making behavior in their presence. Nonetheless, high-performance neural networks like ResNet50 \citep{he2016deep} are easily confused by small local deformations \citep{NIPS2018_7982}.  The juxtaposition of adversarial examples and common corruptions was also explored in \citep{ford2019adversarial} where the authors discuss the relationship between both and encourage researchers working in the field of adversarial robustness to cross-evaluate the robustness of their models towards common corruptions.
    
    We argue that in many practical applications robustness to common corruptions is often more relevant than robustness to artificially designed adversarial perturbations. Autonomous cars should not change their behavior in the face of unusual weather conditions such as hail or sand storms or small pixel defects in their sensors. Not-Safe-For-Work filters should not fail on images with unusual compression artifacts. Likewise, speech recognition algorithms should perform well regardless of the background music or sounds.
    
    Besides its practical relevance, robustness to common corruptions is also an excellent target in its own right for researchers in the field of adversarial robustness and domain adaptation. Common corruptions can be seen as distributional shifts or as a weak form of adversarial examples that live in a smaller, constrained subspace.

    Despite their importance, common corruptions have received relatively little attention so far. Only recently a modification of the ImageNet dataset \citep{DBLP:journals/corr/RussakovskyDSKSMHKKBBF14} to benchmark model robustness against common corruptions and perturbations has been published \citep{hendrycks2018benchmarking} and is referred to as ImageNet-C. Now, this scheme has also been applied to other common datasets resulting in Pascal-C, Coco-C and Cityscapes-C \citep{michaelis2019benchmarking} and MNIST-C \citep{mu2019mnist}.
    
    Our contributions are as follows:
    \begin{itemize}
        \item We demonstrate that data augmentation with Gaussian or Speckle noise serves as a simple yet very strong baseline that is sufficient to surpass almost all previously proposed defenses against common corruptions on ImageNet-C for ResNet50. We further show that the magnitude of the additive noise is a crucial hyper-parameter to reach optimal robustness.
        \item Motivated by our strong results with baseline noise augmentations, we introduce a neural network-based \textit{adversarial noise generator} that can learn arbitrary uncorrelated noise distributions that maximally fool a given recognition network when added to their inputs. We denote the resulting noise patterns as \textit{adversarial noise}.
       \item We design and validate a constrained Adversarial Noise Training (ANT) scheme through which the recognition network learns to become robust against adversarial noise. We demonstrate that our ANT reaches state-of-the-art robustness on the corruption benchmark ImageNet-C for the commonly used ResNet50 architecture and on MNIST-C, even surpassing the already strong baseline noise augmentations. This result is not due to overfitting on the Noise categories of the respective benchmarks since we find equivalent results on the non-noise corruptions as well.
        \item We modify the adversarial noise generator to allow it to produce locally correlated noise thereby enabling it to learn more diverse noise distributions. Performing ANT with the modified noise generator, we observe an increase in robustness for the `snow' corruption which is visually similar to our learned noise.
       \item We demonstrate a further increase in robustness when combining ANT with previous defense methods.
       \item We substantiate the claim that increased robustness against regular or universal adversarial perturbations does not imply increased robustness against common corruptions. This is not necessarily true vice-versa: Our noise trained recognition network has high accuracy on ImageNet-C and also slightly improved accuracy on adversarial attacks on clean ImageNet compared to a vanilla trained ResNet50. 
    \end{itemize}
We released our trained model weights along with evaluation code on \href{https://github.com/bethgelab/game-of-noise}{github.com/bethgelab/game-of-noise}.
    
    \begin{figure*}
      \includegraphics[width=\linewidth]{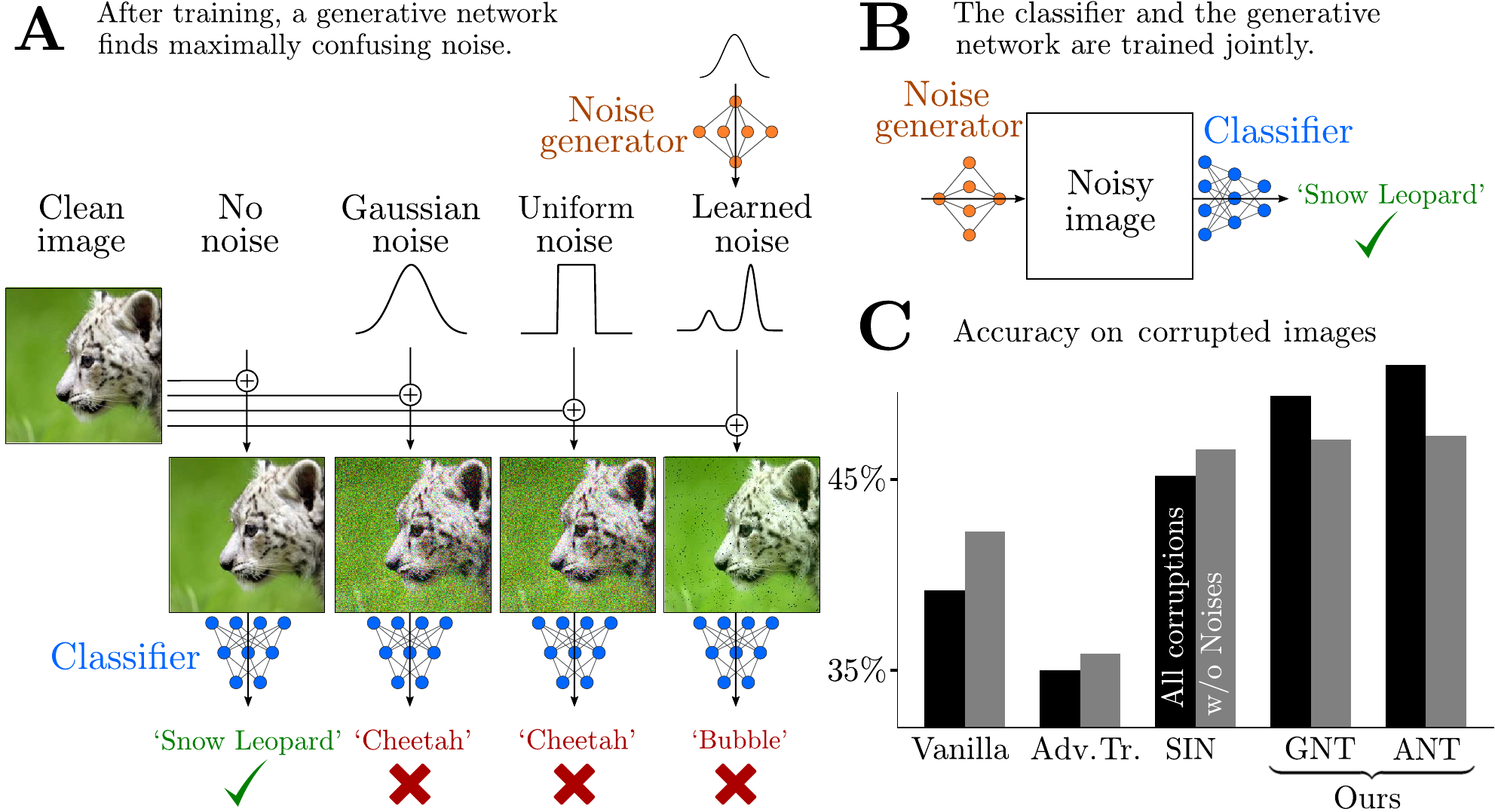}
      \caption{Outline of our approach. A: First, we train a generative network against a vanilla trained classifier to find the adversarial noise. B: To achieve robustness against adversarial noise, we train the classifier and the noise generator jointly. C: We measure the robustness against common corruptions for a vanilla, adversarially trained (Adv. Tr.), trained on Stylized ImageNet (SIN), trained via Gaussian data augmentation (GNT) and trained with the means of Adversarial Noise Training (ANT). With our methods, we achieve the highest accuracy on common corruptions, both on all and non-Noise categories.}
      \label{fig:fig1}
    \end{figure*}

\section{Related work}
\label{sec:relwork}
\paragraph{Robustness against common corruptions}
    Several recent publications study the vulnerability of DNNs to common corruptions. \citet{DBLP:journals/corr/DodgeK16}\ find that state-of-the-art image recognition networks are particularly vulnerable to blur and Gaussian noise. Two recent studies compare humans and DNNs on recognizing corrupted images, showing that DNN performance drops much faster than human performance for increased perturbation sizes \citep{DBLP:journals/corr/DodgeK17b, NIPS2018_7982}. \citet{DBLP:journals/corr/abs-1906-08988}\ study the Fourier properties of common corruptions and link them to the robustness of differently trained classifiers.
    
    \citet{hendrycks2018benchmarking}\ introduce corrupted versions of standard datasets denoted as ImageNet-C, Tiny ImageNet-C and CIFAR10-C as standardized benchmarks for machine learning models and show that while state-of-the-art networks like ResNet50 are more accurate than outdated ones like AlexNet, their robustness is still negligible compared to humans. Similarly, common corruptions have been applied to and evaluated on COCO-C, Pascal-C, Cityscapes-C \citep{michaelis2019benchmarking} and MNIST-C \citep{mu2019mnist}.
    
    There have been attempts to increase robustness against common corruptions. \citet{zhang2019making} integrate an anti-aliasing module from the signal processing domain in the ResNet50 architecture to restore the shift-equivariance which can get lost in deep CNNs. This results both in increased accuracy on clean data and increased generalization to corrupted image samples.
    Concurrent work to ours demonstrates that having more training data \citep{xie2019selftraining, mahajan2018exploring} or using stronger backbones \citep{xie2019selftraining, michaelis2019benchmarking} can significantly improve model performance on common corruptions.
    
    A popular method to decrease overfitting and help the network generalize better to unseen data is to augment the training dataset by applying a set of (randomized) manipulations to the images \citep{Mikoajczyk2018DataAF}. Furthermore, augmentation methods have also been applied to make the models more robust against image corruptions \citep{geirhos2018imagenettrained}. 
     \citet{NIPS2018_7982} train ImageNet classifiers against a fixed set of corruptions but find no generalized robustness against unseen corruptions. However, they considered vastly higher noise severities than us.  A similar observation is made by \citep{DBLP:journals/corr/DodgeK17a}. In a follow-up study, \citet{geirhos2018imagenettrained} show that recognition models are biased towards texture and suggest this bias as one source of susceptibility for corruptions. They demonstrate that an increased shape bias also leads to increased accuracy on corrupted images.
     \citet{hendrycks2020augmix} is concurrent work to ours where the authors propose a data augmentation strategy which relies on combining and mixing augmentation chains. They also report strong robustness increases on ImageNet-C.

    Augmentation with Gaussian noise has been used as a regularizer for smoothing the decision boundary of the classifier and was shown to be a provable adversarial defense \citep{pmlr-v97-cohen19c}.
    Conceptually, \citet{ford2019adversarial} is the closest study to our work, since they also apply Gaussian noise to images to increase corruption robustness. They observe a low relative improvement in accuracy on corrupted images whereas we were able to outperform almost all previous baselines on the commonly used ResNet50 architecture.\footnote{To compare with \citet{ford2019adversarial}, we evaluate our approach for an InceptionV3 architecture, see our results in Appendix~H.} They use a different architecture (InceptionV3 versus our ResNet50) and train a new model from scratch whereas we fine-tune a pretrained model. Another methodological difference is that we split every batch evenly in clean data and data augmented by Gaussian noise whereas they sample the standard deviation uniformly between 0 and one specific value and add noise to each image.
     \citet{Lopes_Gaussian_Patch} restrict the Gaussian noise to small image patches which improves accuracy but does not yield state-of-the-art performance on the ResNet50 architecture.
     
    An alternative approach to data augmentation relies on stability training and shows robustness towards a range of distortions \citep{laermann2019achieving}.

\paragraph{Link between adversarial robustness and common corruptions}
    There is currently no agreement on whether adversarial training increases robustness against common corruptions in the literature.
      \citet{hendrycks2018benchmarking} report a robustness increase on common corruptions due to adversarial logit pairing on Tiny ImageNet-C.
     \citet{ford2019adversarial} suggest a link between adversarial robustness and robustness against common corruptions, claim that increasing one robustness type should simultaneously increase the other, but report mixed results on MNIST and CIFAR10-C. Additionally, they also observe large drops in accuracy for adversarially trained networks and networks trained with Gaussian data augmentation compared to a vanilla classifier on certain corruptions. They do not evaluate adversarially robust classifiers on ImageNet.
    \citet{DBLP:journals/corr/FawziMF16} show that curvature constraints can both improve robustness against adversarial and random perturbations but they only present results on vanilla networks.
    On the other hand, \citet{engstrom2017rotation} report that increasing robustness against adversarial $\ell_\infty$ attacks does not increase robustness against translations and rotations, but they do not present results on noise.
    \citet{kang2019transfer} study robustness transfer between models trained against $\ell_1$, $\ell_2$, $\ell_\infty$ adversaries / elastic deformations and JPEG artifacts. They observe that adversarial training increases robustness against elastic and JPEG corruptions on a 100-class subset of ImageNet. This result contradicts our findings on full ImageNet as we see a slight decline in accuracy on those two classes for the adversarially trained model from \citep{featdenoise} and severe drops in accuracy on other corruptions. 
     \citet{jordan2019quantifying} show that adversarial robustness does not transfer easily between attack classes.
     \citet{Tramer2019Adversarial} also argue in favor of a trade-off between different robustness types. For a simple and natural classification task, they prove that adversarial robustness towards $l_\infty$ perturbations does neither transfer to $l_1$ nor to input rotations and translations, and vice versa. They support their formal analysis with experiments on MNIST and CIFAR10.

\paragraph{Universal adversarial perturbations}
    Universal adversarial perturbations (UAPs) \citep{DBLP:journals/corr/Moosavi-Dezfooli16} are perturbations which, if added to any image, fool a given recognition model. This contrasts with regular adversarial perturbations, which need to be designed specifically for every single image. \citet{DBLP:journals/corr/abs-1708-05207} generate UAPs by training so-called universal adversarial networks (UANs). They also train the classifier jointly with the UAN but manage to only slightly increase robustness against UAPs. Other defenses against UAPs are similarly based on adversarial training \citep{hendrik2018universality,DBLP:journals/corr/abs-1811-11304,DBLP:journals/corr/abs-1812-03705,DBLP:journals/corr/abs-1809-07802}.
    
    UAPs are very different from our adversarial noise setting in that UAPs can learn perturbations with global, image-wide features while our adversarial noise is identically distributed over pixels and thus inherently local.

\section{Methods}
    Our method section can be roughly split into two parts. In the first part, we revisit Gaussian data augmentation as a method to increase robustness against image corruptions. The second part contains two items (Fig.~\ref{fig:fig1}). First, we devise and train a generator neural network to produce spatially uncorrelated noise that is adversarial to a given recognition network (section~\ref{sub:adversarialnoise} and Fig.~\ref{fig:fig1}A). Second, we formulate a constrained adversarial training scheme, which allows us to train the recognition model jointly with the noise generator (section~\ref{sub:jointtraining} and Fig.~\ref{fig:fig1}B) that we call Adversarial Noise Training (ANT). Finally, in section~\ref{sub:evaluation} we explain our evaluation methods on corrupted images (Fig.~\ref{fig:fig1}C).

\subsection{Training with Gaussian noise}
    As discussed in section \ref{sec:relwork}, several researchers have tried using Gaussian noise as a method to increase robustness towards common corruptions with mixed results.
    In this work, we revisit the approach of Gaussian data augmentation and increase its efficacy. In contrast to previous work, we treat the standard deviation $\sigma$ of the distribution as a hyper-parameter of the training and measure its influence on robustness. 
    
    To formally introduce the objective, let $\mathcal{D}$ be the data distribution over input pairs $(\x, y)$ with $\x\in\mathbb{R}^N$ and \mbox{$y \in \{1,..., k\}$.} 
    We train a differentiable classifier $f_\theta(\x)$ by minimizing the risk on a dataset with additive Gaussian noise 
    \begin{equation}
        \E_{\x, y \sim \mathcal{D}} \, \E_{\bdelta \sim \mathcal{N}(\mathbf{0}, \sigma^2 \mathds{1})} \left[ \mathcal{L}_{\CE}\left(f_\theta(\operatorname{clip}((\x + \bdelta)), y\right)\right],
    \end{equation}
    where $\sigma$ is the standard deviation of the Gaussian noise and $\x + \bdelta$ is clipped to the input range $[0, 1]^N$. The standard deviation is either kept fixed or is chosen uniformly from a fixed set of standard deviations. In both cases, the possible standard deviations are chosen from a small set of nine values inspired by the noise variance in the ImageNet-C dataset (cf. section \ref{sub:evaluation}). 
    To maintain high accuracy on clean data, we only perturb 50\% of the training data with Gaussian noise within each batch.
    
\subsection{Adversarial noise}
\label{sub:advnoise}

\subsubsection{Learning Adversarial Noise}
\label{sub:adversarialnoise}

    Our goal is to find a noise distribution $p_\phi(\bdelta),\, \bdelta\in\mathbb{R}^N$ such that noise samples added to $\x$ maximally confuse the classifier $f_\theta$. More concisely, we optimize
    \begin{equation}
        \label{eq:generatorupdate}
        \begin{split}
            \max_\phi \E_{\x, y \sim \mathcal{D}} \, \E_{\bdelta \sim p_\phi(\bdelta)}\left[\mathcal{L}_{\CE}\left(f_\theta(\operatorname{clip}_{\epsilon}(\x + \bdelta)), y\right)\right],
        \end{split}
    \end{equation}
    where $\operatorname{clip}$ is an operator that clips all values to the valid interval (i.e. $\operatorname{clip}(\x + \bdelta) \in [0, 1]^N$) and $||\bdelta||_2 = \epsilon$. \footnote{We apply the method derived in \cite{rauber2020fast} and rescale the perturbation by a factor $\gamma$ to obtain the desired $\ell_2$ norm; despite the clipping, the squared $\ell_2$ norm is a piece-wise linear function of $\gamma^2$ that can be inverted to find the correct scaling factor $\gamma$.}
    
    We do not have to explicitly model the probability density function $p_\phi(\bdelta)$ since optimizing Eq.~\eqref{eq:generatorupdate} only involves samples drawn from $p_\phi(\bdelta)$. We model the samples from $p_\phi(\bdelta)$ as the output of a neural network $g_\phi : \mathbb{R}^N \rightarrow \mathbb{R}^N$ which gets its input from a normal distribution $\bdelta = g_\phi(\z)$ where $\z \sim \mathcal{N}(\mathbf{0}, \mathds{1})$. We enforce the independence property of $p_\phi(\bdelta) = \prod_{n}p_\phi(\delta_n)$ by constraining the network architecture of the noise generator $g_\phi$ to only consist of convolutions with 1x1 kernels. Lastly, the projection onto a sphere  $||\bdelta||_2 = \epsilon$ is achieved by scaling the generator output with a scalar while clipping $\x + \bdelta$ to the valid range $[0, 1]^N$. This fixed size projection (hyper-parameter) is motivated by the fact that Gaussian noise training with a single, fixed $\sigma$ achieved the highest accuracy. \footnote{We also experimented with an adaptive sphere radius $\epsilon$ which grows with the classifier's accuracy. However, we did not see any improvements and followed Occam's razor.}

    The noise generator $g_\phi$ has four 1x1 convolutional layers with ReLU activations and one residual connection from input to output. The convolutional weights are initialized such that the noise generator outputs a Gaussian distribution.

\subsubsection{Adversarial Noise Training}
\label{sub:jointtraining}

    To increase robustness, we now train the classifier $f_\theta$ to minimize the risk under adversarial noise distributions jointly with the noise generator
    \begin{equation}
        \label{eq:classifierobjective}
        \begin{split}
            \min_\theta \max_\phi \E_{\x, y \sim \mathcal{D}} \, \E_{\bdelta \sim p_\phi(\bdelta)}\left[\mathcal{L}_{\CE}\left(f_\theta(\operatorname{clip}_{\epsilon}(\x + \bdelta)), y\right)\right],
        \end{split}
    \end{equation}
    where again $\x + \bdelta \in [0, 1]^N$ and $||\bdelta||_2 = \epsilon$.
    For a joint adversarial training, we alternate between an outer loop of classifier update steps and an inner loop of generator update steps. This is also depicted schematically in Fig.~\ref{fig:fig1}B. Note that in regular adversarial training, e.g.\ \citep{madry2018towards}, $\bdelta$ is optimized directly whereas we optimize a constrained distribution over $\bdelta$.

    To maintain high classification accuracy on clean samples, we sample every mini-batch so that they contain $50\%$ clean data and perturb the rest. The current state of the noise generator is used to perturb $30\%$ of this data and the remaining $20\%$ are augmented with samples chosen randomly from previous distributions. For this, the noise generator states are saved at regular intervals.
    The latter method is inspired by experience replay from reinforcement learning \citep{mnih2015human} and is used to keep the classifier from forgetting previous adversarial noise patterns.
    
    To prevent the noise generator from being stuck in a local minimum, we halt the Adversarial Noise Training (ANT) at regular intervals and train a new noise generator from scratch. This noise generator is trained against the current state of the classifier to find a current optimum. The new noise generator replaces the former noise generator in the ANT. This technique has proven crucial to train a robust classifier.

\paragraph{Learning locally correlated adversarial noise}
\label{sub:ng_3x3}
We modify the architecture of the noise generator defined in Eq. \ref{eq:generatorupdate} to allow for local spatial correlations and thereby enable the generator to learn more diverse distributions. Since we seek to increase model robustness towards image corruptions such as rain or snow that produce locally correlated patterns, it is natural to include local patterns in the manifold of learnable distributions. We replace the 1x1 kernels in one network layer with 3x3 kernels limiting the maximum correlation length of the output noise sample to 3x3 pixels. 
We indicate the correlation length of noise generator used for the constrained adversarial noise training as ANT$^{1\mathrm{x}1}$ or ANT$^{3\mathrm{x}3}$.

\subsection{Combining Adversarial Noise Training with stylization}
    As demonstrated by \citet{geirhos2018imagenettrained}, using random stylization as data augmentation increases the accuracy on ImageNet-C. 
    The robustness gains are attributed to a stronger shape bias of the classifier. We combine our ANT and the stylization approach to achieve robustness gains from both.
    To incorporate stylized data into the training scheme described in the previous section, we change the way we sample every mini-batch: we split the batches into clean data (25\%), stylized data (30\%) and data perturbed by the noise generator (45\%). Like before, we choose the latter samples with roughly equal probability from the current state of the noise generator and from any previous distribution.

\subsection{Evaluation on corrupted images}
\label{sub:evaluation}

\paragraph{Evaluation of noise robustness}
    We evaluate the robustness of a model by sampling a Gaussian noise vector $\bdelta$. We then do a line search along the direction $\bdelta$ starting from the original image $\x$ until it is misclassified. We denote the resulting minimal perturbation as $\bdelta_{\text{min}}$. The robustness of a model is then denoted by the median\footnote{Samples for which no $\ell_2$-distance allows us to manipulate the classifier's decision contribute a value of $\infty$ to the median.} over the test set

    \begin{equation}
        \epsilon^* = \median_{\x, y \sim \mathcal{D}}\,||\bdelta_\text{min}||_2,
        \label{eq:eval}      
    \end{equation}
    with $f_\theta(\x + \bdelta_\text{min}) \ne y$ and $\x + \bdelta_\text{min} \in [0, 1]^N$.

    Note that a higher $\epsilon^*$ denotes a more robust classifier.
    To test the robustness against adversarial noise, we train a new noise generator at the end of the Adversarial Noise Training until convergence and evaluate it according to Eq.~\eqref{eq:eval}.

\paragraph{ImageNet-C}
    The ImageNet-C benchmark\footnote{For the evaluation, we use the JPEG compressed images from \href{https://github.com/hendrycks/robustness}{github.com/hendrycks/robustness} as is advised by the authors to ensure reproducibility. We note that \citet{ford2019adversarial} report a decrease in performance when the compressed JPEG files are used as opposed to applying the corruptions directly in memory without compression artifacts.} \citep{hendrycks2018benchmarking} is a conglomerate of 15 diverse corruption types that were applied to the validation set of ImageNet.
    The corruptions are organized into four main categories: Noise, Blur, Weather, and Digital and have five levels of severities to reflect the varying intensities of common corruptions.
    The MNIST-C benchmark is created similarly to ImageNet-C \citep{mu2019mnist} with a slightly different set of corruptions.
    Our main evaluation metric for both benchmarks is the Top-1 accuracy on corrupted images for each Noise category averaged over the severities; we also report the Top-5 accuracy on ImageNet-C. Since some works report the `mean Corruption Error' (mCE) instead of accuracy, we also include results on mCE in Appendix~D.
    
    We evaluate all proposed methods for ImageNet-C on the ResNet50 architecture for better comparability to previous methods, e.g.\ \citep{geirhos2018imagenettrained, Lopes_Gaussian_Patch, zhang2019making}. The clean ImageNet accuracy of the used architecture highly influences the results and could be seen as an upper bound for the accuracy on ImageNet-C. Note that our approach is independent of the used architecture and could in principle be applied to any differentiable network. 

\section{Results}
    For our experiments on ImageNet, we use a classifier that was pretrained on ImageNet. For the experiments on MNIST, we use the architecture from \citep{madry2018towards} for comparability. We use PyTorch \citep{paszke2017automatic} for all of our experiments.
    All technical details, hyper-parameters and the architecture of the noise generator can be found in Appendix~A-B.
    
\subsection{(In-)Effectiveness of regular adversarial training to increase robustness towards common corruptions}    
\label{sub:regular_adv_training}
    As our first experiment, we evaluate whether robustness against regular adversarial examples generalizes to robustness against common corruptions.
    We display the Top-1 accuracy of vanilla and adversarially trained models in Table~\ref{tab_adversarial_training_cc}; detailed results on individual corruptions can be found in Appendix~C.
    For all tested models, we find that regular $\ell_{\infty}$ adversarial training can strongly decrease the robustness towards common corruptions, especially for the corruption types Fog and Contrast. Universal adversarial training \citep{DBLP:journals/corr/abs-1811-11304}, on the other hand, leads to severe drops on some corruptions but the overall accuracy on ImageNet-C is slightly increased relative to the vanilla baseline model (AlexNet). Nonetheless, the absolute ImageNet-C accuracy of 22.2\% is still very low. These results disagree with two previous studies which reported that (1) adversarial logit pairing\footnote{Note that ALP was later found to not increase adversarial robustness \citep{logitpairing2018}.} (ALP) increases robustness against common corruptions on Tiny ImageNet-C \citep{hendrycks2018benchmarking}, and that (2) adversarial training can increase robustness on the CIFAR10-C data set \citep{ford2019adversarial}.
   
    We evaluate adversarially trained models on MNIST-C and present the results and their discussion in Appendix~E. The results on MNIST-C show the same tendency as on ImageNet-C: adversarially trained models have lower accuracy on MNIST-C and thus indicate that adversarial robustness does not transfer to robustness against common corruptions. This corroborates the results of \citep{ford2019adversarial} on MNIST who also found that an adversarially robust model had decreased robustness towards a set of common corruptions.

    \begin{table}[h]
    \begin{center}
      \begin{tabular}{l c c} 
    
        model              & IN-C & IN-C w/o Noises \\
                    \\[-1em]
        \hline 
                    \\[-1em]
        Vanilla RN50                    & 39.2\%   & 42.3\% \\
        Adv. Training \citep{adv_training_free}   & 29.1\%  & 32.0\%\\
                    \\[-1em]
        \hline
                    \\[-1em]
        Vanilla RN152                   & 45.0\%   & 47.9\% \\
        Adv. Training \citep{featdenoise}               & 35.0\%   & 35.9\%  \\  
                    \\[-1em]
        \hline
                    \\[-1em]
        Vanilla AlexNet       & 21.1\% &  23.9\%\\
        Universal Adv. Training \citep{DBLP:journals/corr/abs-1811-11304} & 22.2\% & 23.1\%\\
      \end{tabular}
    \end{center}
    \caption{Top-1 accuracy on ImageNet-C and ImageNet-C without the Noise categories (higher is better). Regular adversarial training decreases robustness towards common corruptions; universal adversarial training seems to slightly increase it.}
    \label{tab_adversarial_training_cc}
    \end{table}

\subsection{Effectiveness of Gaussian data augmentation to increase robustness towards common corruptions}
\label{sub:gaussiandata}
\begin{figure*}
    \begin{center}
      \includegraphics[width=1.0\textwidth]{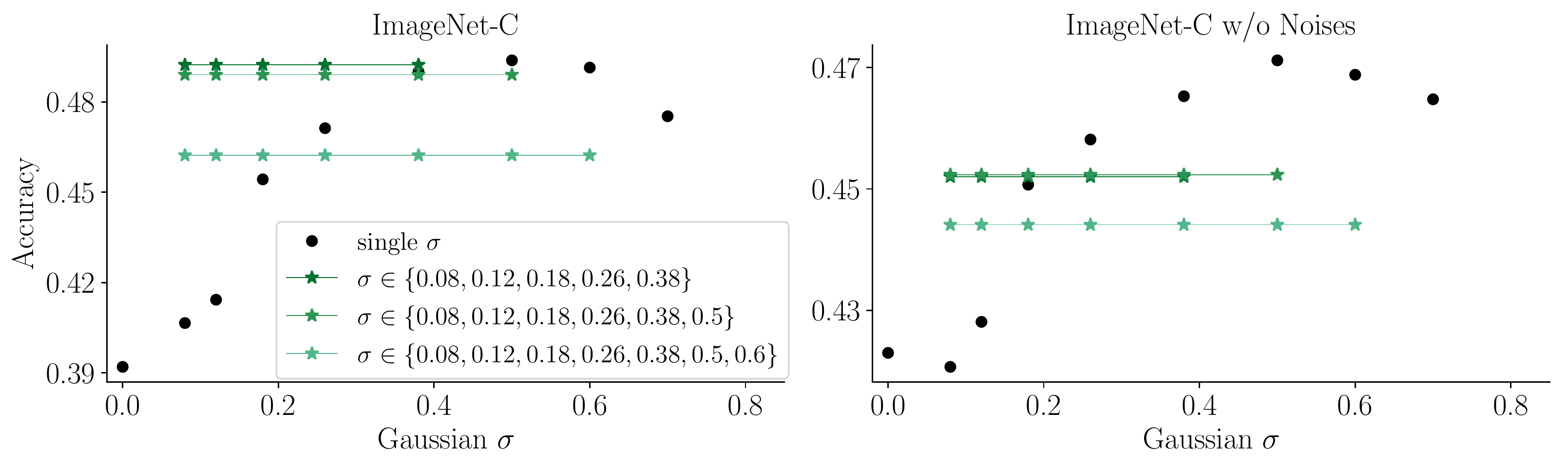}
      \caption{Top-1 accuracy on ImageNet-C (left) and ImageNet-C without the noise corruptions (right) of a ResNet50 architecture fine-tuned with Gaussian data augmentation of varying standard deviation $\sigma$. We train on Gaussian noise sampled from a distribution with a single $\sigma$ (black dots) and on distributions where $\sigma$ is sampled from different sets (green lines with stars). We also compare to a vanilla trained model at $\sigma=0$.}
            \label{fig:fig_gauss}
      \end{center}
    \end{figure*}

    We fine-tune a pretrained image classifier with Gaussian data augmentation from the distribution $\mathcal{N}(\mathbf{0}, \sigma^2 \mathds{1})$ and vary $\sigma$. We try two different settings: in one we choose a single noise level $\sigma$ while in the second we sample $\sigma$ uniformly from a set of multiple possible values. The Top-1 accuracy of the fine-tuned models on ImageNet-C and a comparison to a vanilla trained model is shown in Fig.~\ref{fig:fig_gauss}.
    Each black point shows the performance of one model fine-tuned with one specific $\sigma$; the vanilla trained model is marked by the point at $\sigma=0$.
    The horizontal lines indicate that the model is fine-tuned with Gaussian noise where $\sigma$ is sampled from a set for each image. For example, for the dark green line, as indicated by the stars, we sample $\sigma$ from the set $\{0.08, 0.12, 0.18, 0.26, 0.38 \}$, which corresponds to the Gaussian corruption of ImageNet-C. We show both the results on the full ImageNet-C evaluation set and the results on ImageNet-C without Noises (namely Blur, Weather and Digital) since Gaussian noise is part of the test set. To give a feeling of how the different $\sigma$-levels manifest themselves in an image, we include example images for all $\sigma$-levels in Appendix~G. 

    There are three important results evident from Fig.~\ref{fig:fig_gauss}: 
    \begin{enumerate}
        \item  Gaussian noise generalizes well to the non-noise corruptions of the ImageNet-C evaluation dataset and is a powerful baseline. This is a surprising result as it was shown in several recent works that training on Gaussian or uniform noise does not generalize to other corruption types \citep{NIPS2018_7982, Lopes_Gaussian_Patch} or that the effect is very weak \citep{ford2019adversarial}. 

        \item  The standard deviation $\sigma$ is a crucial hyper-parameter and has an optimal value of about $\sigma=0.5$ for ResNet50.
        \item If $\sigma$ is chosen well, using a single $\sigma$ is enough, sampling from a set of $\sigma$ values is not necessary and even detrimental for robustness against non-noise corruptions.
    \end{enumerate}
    
    In the following Results sections, we will compare Gaussian data augmentation to our Adversarial Noise Training approach and baselines from the literature. For this, we will use the models with the overall best-performance: The model GN$_{0.5}$ that was trained with Gaussian data augmentation with a single $\sigma=0.5$ and the model GN$_{\mathrm{mult}}$ where $\sigma$ was sampled from the set $\{0.08, 0.12, 0.18, 0.26, 0.38\}$.

\subsection{Evaluation of the severity of adversarial noise as an attack}
    
    In this section, we try to answer the question: Can we learn the most severe uncorrelated additive noise distribution for a classifier?
    
    Following the success of simple uncorrelated Gaussian noise data augmentation (section~\ref{sub:gaussiandata}) and the ineffectiveness of regular adversarial training (section~\ref{sub:regular_adv_training}) which allows for highly correlated patterns, we restrict our learned noise distribution to be sampled independently for each pixel. We denote this learned adversarial noise distribution $p_\phi(\bdelta)$ as adversarial noise (AN, section~\ref{sub:adversarialnoise}).
    
    To measure the effectiveness of our adversarial noise, we report the median perturbation size $\epsilon^*$ that is necessary for a misclassification for each image in the test set. 
   
    In Table~\ref{tab:epssev}, we see that our AN is much more effective at fooling the classifier compared to Gaussian and uniform noise. This is also reflected qualitatively in the noisy images in Fig.~\ref{fig:fig1} where we show images at the decision boundary: The amount of noise to fool the classifier is smaller in the right-most image produced by the generative network than in the central images (Gaussian and uniform noise). 
  
    \begin{table}[ht]
      \begin{center}
          \begin{tabular}{l c c c}
            model  & $\epsilon^*_{\mathrm{GN}}$ & $\epsilon^*_{\mathrm{UN}}$ &  $\epsilon^*_{\mathrm{AN}}$\\
            \hline 
            Vanilla RN50 & 39.0 & 39.1 & 16.2  \\  
          \end{tabular}
          \caption{Median $\ell_2$ perturbation size $\epsilon^*$ that is required to misclassify an image for Gaussian (GN), uniform (UN) and adversarial noise (AN). A lower $\epsilon^*$ indicates a more severe noise, since on average, a smaller perturbation size is sufficient to fool a classifier.}
         \label{tab:epssev}
      \end{center}
    \end{table}

\subsection{Evaluation of Adversarial Noise Training as a defense}
 In the previous section, we established a method for learning the most adversarial noise distribution for a classifier. Now, we utilize it for a joint Adversarial Noise Training (ANT$^{1\mathrm{x}1}$) where we simultaneously train the noise generator and classifier (see section \ref{sub:jointtraining}). 
    This leads to substantially increased robustness against Gaussian, uniform and adversarial noise, see Table \ref{tab:jt}. 
   
    The robustness of models that were trained via Gaussian data augmentation also increases, but on average much less compared to the model trained with ANT$^{1\mathrm{x}1}$. To evaluate the robustness against adversarial noise, we train four noise generators from scratch with different random seeds and measure $\epsilon^*_{\mathrm{AN}1\mathrm{x}1}$. We report the mean value and the standard deviation over the four runs.
   
    To better understand this effect, we visualize the temporal evolution of the probability density function $p_\phi(\delta_n)$ of uncorrelated noise during ANT$^{1\mathrm{x}1}$ and samples of correlated noise during ANT$^{3\mathrm{x}3}$ in Fig.~\ref{fig:histograms}. This shows that the generator converges to different distributions during ANT. Therefore, the classifier has been trained against a rich variety of distributions.

    For ANT$^{1\mathrm{x}1}$, we have also tested how the classifier's performance changes if it is trained against adversarial noise sampled randomly from $p_{\phi}(\delta_n)$. The accuracy on ImageNet-C decreases slightly compared to regular ANT$^{1\mathrm{x}1}$: 51.1\%/ 71.9\% (Top-1/ Top-5) on full ImageNet-C and 47.3\%/ 68.3\% (Top-1/ Top-5) on ImageNet-C without the Noise categories. 

    \begin{table}[ht]
      \begin{center}
          \begin{tabular}{l c c c c} 
            model  & clean acc. & $\epsilon^*_{\mathrm{GN}}$ & $\epsilon^*_{\mathrm{UN}}$ &  $\epsilon^*_{\mathrm{AN}1\mathrm{x}1}$\\
            \hline 
            Vanilla RN50 & 76.1\% & 39.0 & 39.1 & 15.7 $\pm$ 0.6  \\ 
            GNT$\sigma_{0.5}$ & 75.9\%  & 74.8   & 74.9    & 31.8 $\pm$ 3.9  \\
            ANT$^{1\mathrm{x}1}$                     & 76.0\%   &   136.7  &   137.0 & 95.4 $\pm$ 5.7 \\
           \end{tabular}
           \vspace{-0.5cm}
          \caption{Accuracy on clean data and robustness of differently trained models as measured by the median perturbation size $\epsilon^*$. A higher $\epsilon^*$ indicates a more robust model.
          To provide an intuition for the perturbation sizes indicated by $\epsilon^*$, we show example images for Gaussian noise below in Fig.~\ref{Figure:ex_tab_3_2} and a larger Figure for different noise types in Appendix I.}
         \label{tab:jt}
      \end{center}
    \end{table}

\begin{figure}
\setlength\tabcolsep{0.0pt} 
\centering
\begin{tabular}{c c c c}

\subfloat{\includegraphics[width=0.22\textwidth]{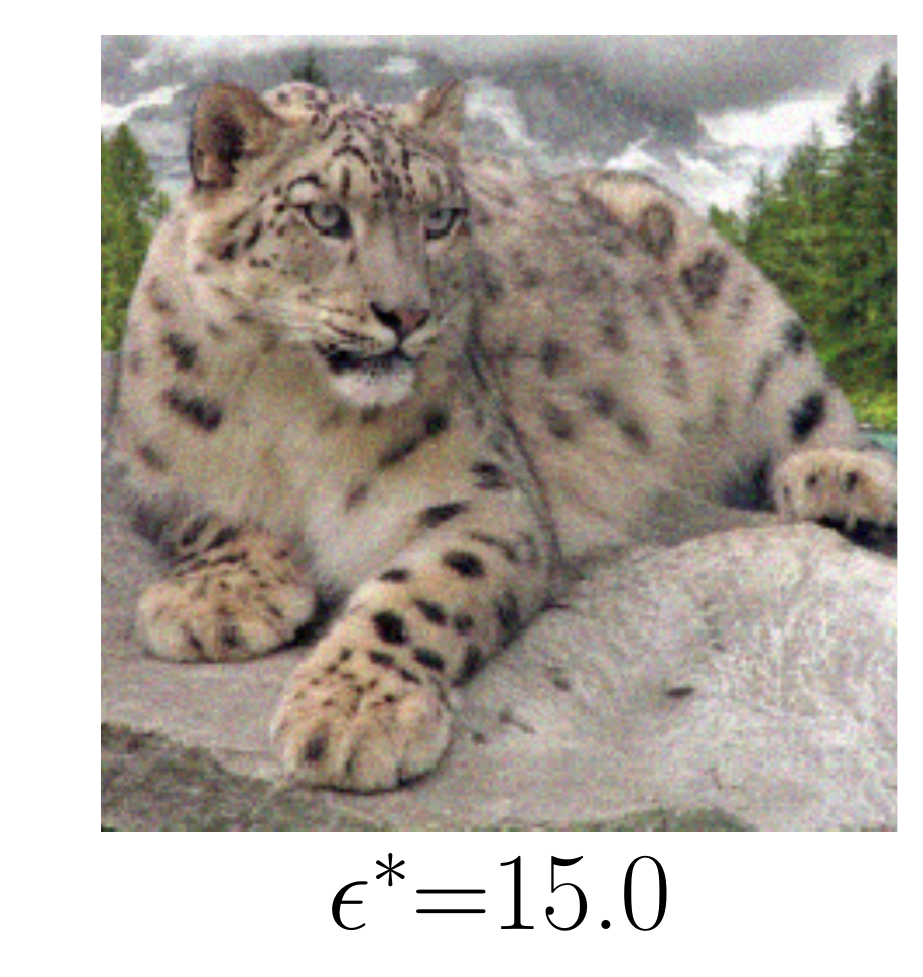}}
& \subfloat{\includegraphics[width=0.22\textwidth]{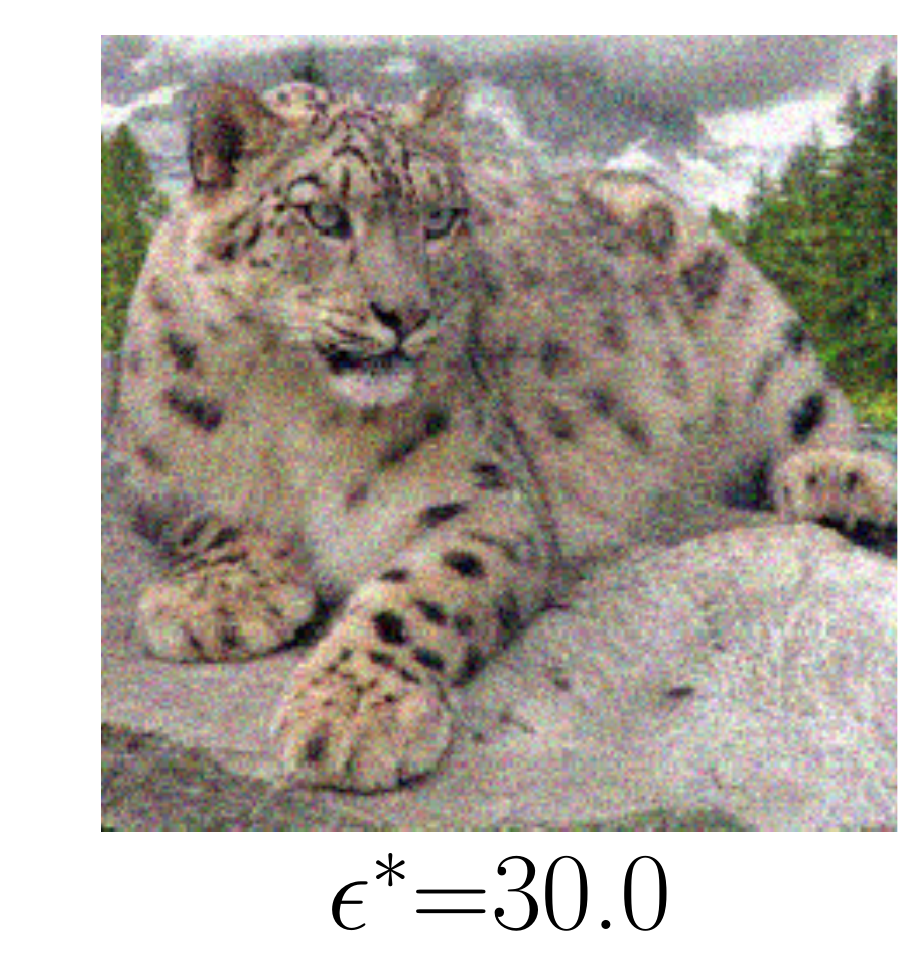}}
& \subfloat{\includegraphics[width=0.22\textwidth]{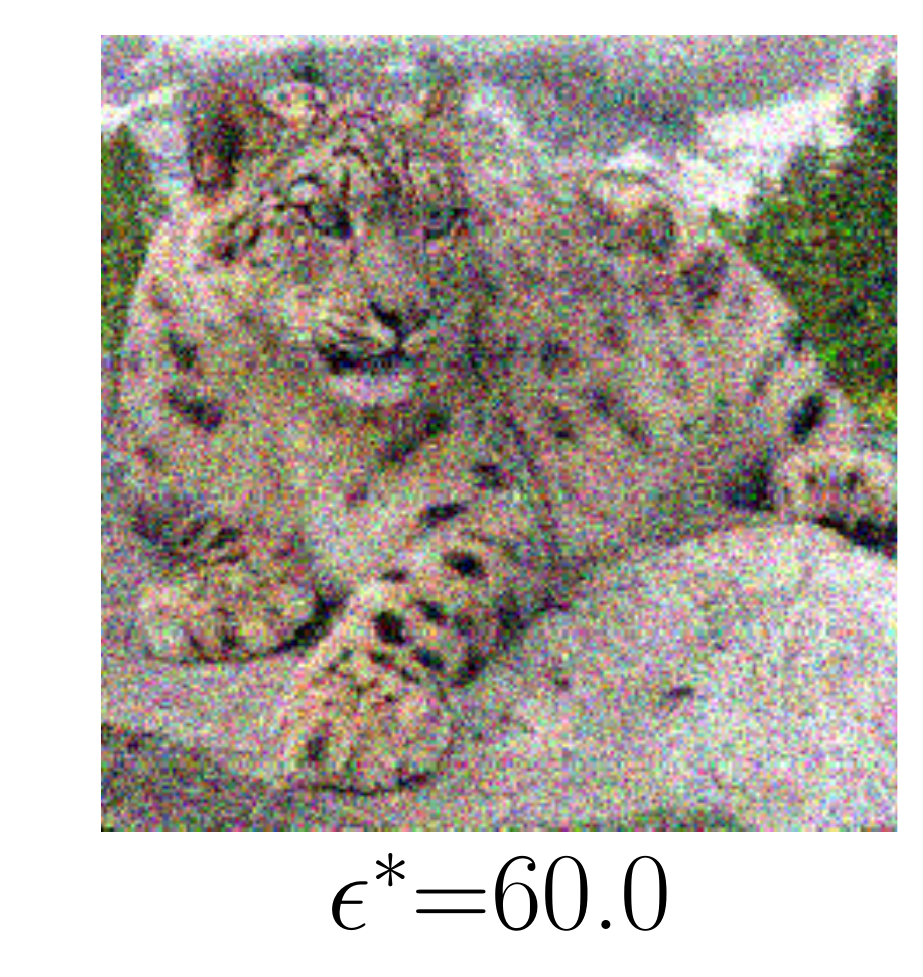}} 
& \subfloat{\includegraphics[width=0.22\textwidth]{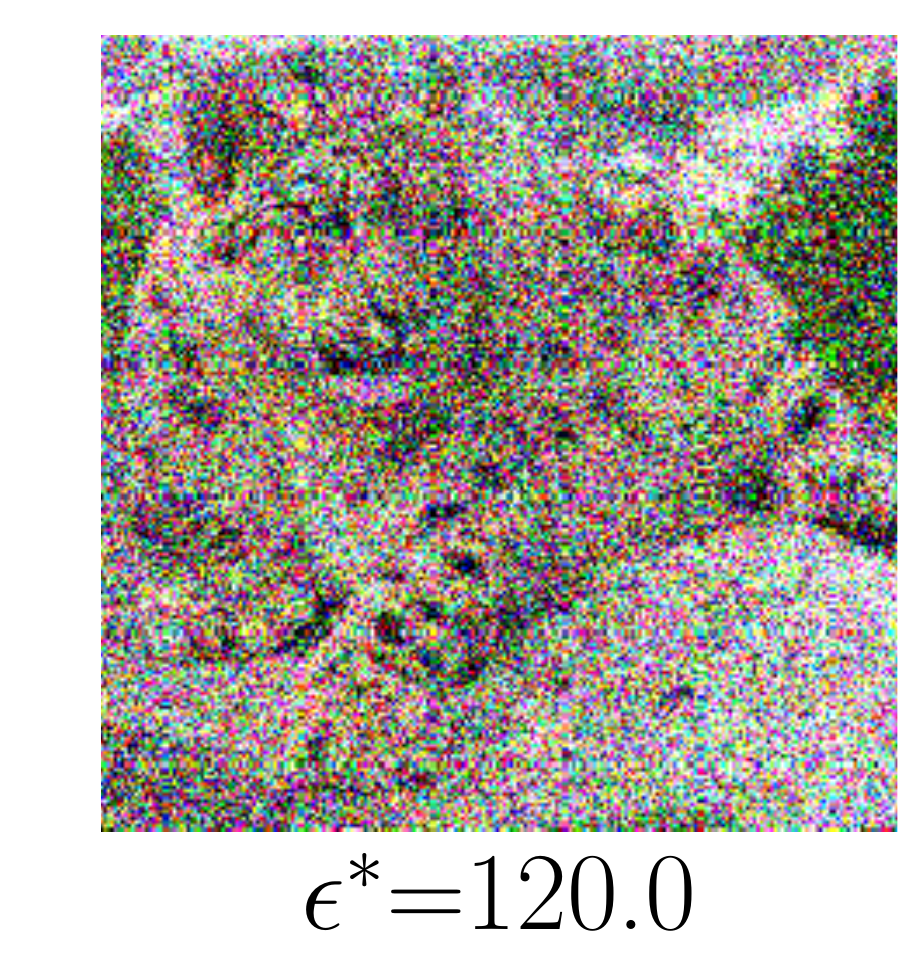}}
\\
\end{tabular}

 \caption{Example images for different perturbation sizes $\epsilon^*$ for additive Gaussian noise on ImageNet.}
 \label{Figure:ex_tab_3_2}
\end{figure}

    \begin{figure}[ht]
    \begin{center}
        \includegraphics[width=\linewidth]{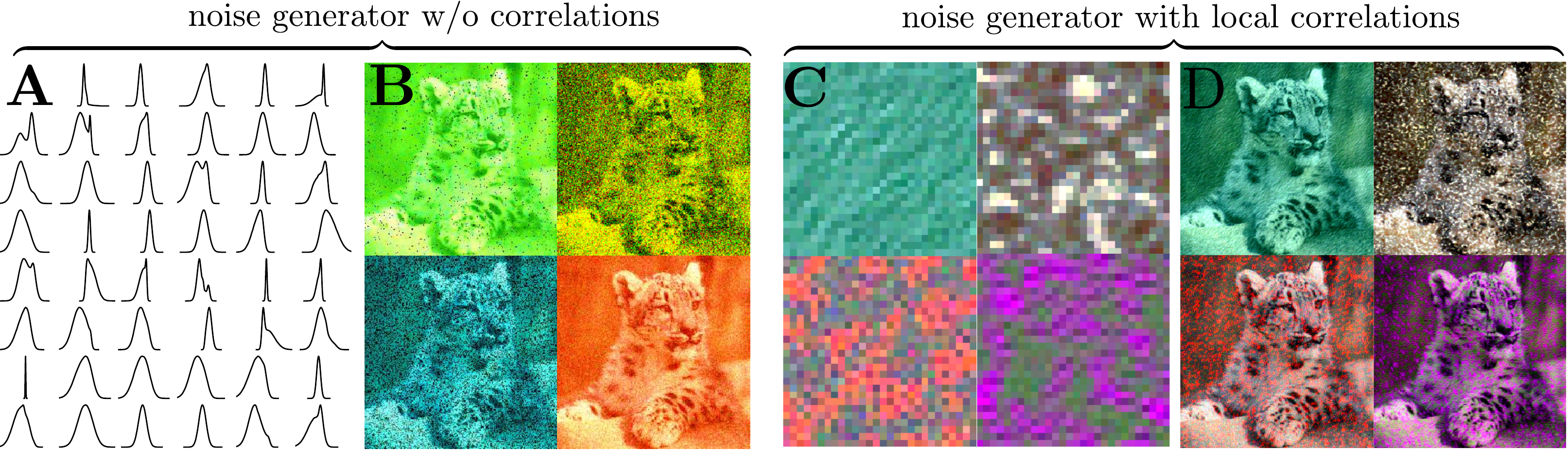}
        \caption{A: Examples of learned probability densities over the grayscale version of the noise $\bdelta_n$ during ANT$^{1\mathrm{x}1}$ where each density corresponds to one local minimum; B: Example images with sampled uncorrelated adversarial noise; C: Example patches of locally correlated noise with a size of 28x28 pixels learned during ANT$^{3\mathrm{x}3}$; D: Example images with sampled correlated adversarial noise.
        }
        \label{fig:histograms}
        \end{center}
    \end{figure}
\vspace{-0.5cm}

\subsection{Comparison of different methods to increase robustness towards common corruptions}

We now revisit common corruptions on ImageNet-C and compare the robustness of differently trained models.
Since Gaussian noise is part of ImageNet-C, we train another baseline model with data augmentation using the Speckle noise corruption from the ImageNet-C holdout set. We later denote the cases where the corruptions present during training are part of the test set by putting corresponding accuracy values in brackets.

The Top-1 accuracies on the full ImageNet-C dataset and ImageNet-C without the Noise corruptions are displayed in Table \ref{tab:IN-C1}; detailed results on individual corruptions in terms of accuracy and mCE are shown in Tables~3 and 4, Appendix~D. We also calculate the accuracy on corruptions without the Noise category as our approaches to add Gaussian noise or to perform ANT$^{1\mathrm{x}1}$ overfits to the Noise categories.
By allowing the noise generator to learn local correlations, we lift the independence constraint and argue that the learned noise distributions are now different enough to not cause overfitting to the Noise categories (see Fig.~\ref{fig:histograms}C+D). For this reason, we do not put the accuracy values of the classifier obtained via ANT$^{3\mathrm{x}3}$ in brackets.

Additionally, we compare our results with several baseline models from the literature. The ImageNet-C benchmark has been published recently and we use all baselines we could find for a ResNet50 architecture:
\begin{enumerate}
    \item Shift Inv: The model is modified to enhance shift-equivariance using anti-aliasing \citep{zhang2019making}.\footnote{Weights were taken from \href{https://github.com/adobe/antialiased-cnns}{github.com/adobe/antialiased-cnns}.}  
    \item Patch GN: The model was trained on Gaussian patches \citep{Lopes_Gaussian_Patch}. Since no model weights are released, we can only include their Top-1 ImageNet-C accuracy values from their paper (and not the Top-5).
    \item SIN+IN: The model was trained on a stylized version of ImageNet \citep{geirhos2018imagenettrained}.\footnote{Weights were taken from \href{https://github.com/rgeirhos/texture-vs-shape/tree/master/models}{github.com/rgeirhos/texture-vs-shape}.}
    \item AugMix: \citet{hendrycks2020augmix} trained their model using diverse augmentations.\footnote{Weights were taken from \href{https://github.com/google-research/augmix}{github.com/google-research/augmix}.} They use image augmentations from AutoAugment \citep{cubuk2018autoaugment} and exclude the contrast, color, brightness, sharpness, and Cutout operations to make sure that the test set of ImageNet-C is disjoint from the training set. We would like to highlight the difficulty in clearly distinguishing between the augmentations used during training and testing as there might be a certain overlap. This can be seen by the visual similarity between the Posterize operation and the JPEG corruption (see Appendix J).
\end{enumerate}

The results on full ImageNet-C are striking (see Table~\ref{tab:IN-C1}): a very simple baseline, namely a model trained with Speckle noise data augmentation, beats almost all previous baselines reaching an accuracy of 46.4\% which is larger than the accuracy of SIN+IN (45.2\%) and close to AugMix (48.3\%).
The GN$\sigma_{0.5}$ surpasses SIN+IN not only on the Noise categories but also on almost all other corruptions, see a more detailed breakdown in Table~\ref{tab:IN-C2}, Appendix~D. The ANT$^{3\mathrm{x}3}$+SIN model produces the best results on ImageNet-C both with and without Noises. Thus, it is slightly superior to Gaussian data augmentation and pure ANT$^{3\mathrm{x}3}$. Comparing ANT$^{1\mathrm{x}1}$ and ANT$^{3\mathrm{x}3}$, we observe that ANT$^{3\mathrm{x}3}$ performs better than ANT$^{1\mathrm{x}1}$ in the  `snow' corruption. We attribute this to the successful modeling capabilities of locally correlated patterns resembling snow of the 3x3 noise generator.
We perform an ablation study to investigate the necessity of experience replay and note that we lose roughly 2\% without it (ANT$^{1\mathrm{x}1}$ w/o EP vs ANT$^{1\mathrm{x}1}$). We also test how the classifier's performance changes if it is trained against adversarial noise sampled randomly from $p_{\phi}(\delta_n)$. The accuracy on ImageNet-C decreases slightly compared to regular ANT$^{1\mathrm{x}1}$: 51.1\%/ 71.9\% (Top-1/ Top-5) on full ImageNet-C and 47.3\%/ 68.3\% (Top-1/ Top-5) on ImageNet-C without the noise category. We include additional results for ANT$^{1\mathrm{x}1}$ with a DenseNet121 architecture \cite{densenet} and for varying parameter counts of the noise generator in Appendix~K.

For MNIST, we train a model with Gaussian data augmentation and via ANT$^{1\mathrm{x}1}$. We achieve similar results with both approaches and report a new state-of-the-art accuracy on MNIST-C: 92.4\%. The results on MNIST-C can be found in Appendix~E.

\begin{table}
  \begin{center}
  \setlength\tabcolsep{0.5pt} 
      \begin{tabular}{l c c c c c} 
         & IN & \multicolumn{2}{c}{IN-C} & \multicolumn{2}{c}{IN-C w/o Noises} \\
            & clean acc. & $\;$ Top-1$\;$ & $\;$Top-5$\;$ &$\;$ Top-1$\;$ &$\;$ Top-5\\
        model & $\;$[\%]$\;$ & $\;$[\%]$\;$ & $\;$[\%]$\;$ & $\;$[\%]$\;$ & $\;$[\%]$\;$ \\
        \hline 
        Vanilla RN50 & 76.1 & 39.2 & 59.3  & 42.3 & 63.2\\
        Shift Inv \citep{zhang2019making} & 77.0  & 41.4 &  61.8   & 44.2&  65.1\\
        Patch GN \citep{Lopes_Gaussian_Patch} & 76.0  & (\g{43.6}) &  (\g{n.a.})    &  43.7& n.a.\ \\
        SIN+IN \citep{geirhos2018imagenettrained}  & 74.6 & 45.2 &  66.6   & 46.6 & 68.2\\
        AugMix \citep{hendrycks2020augmix}  &77.5 & 48.3 & 69.2   & 50.4 & 71.8 \\[0.5em]

        GNT$_{\mathrm{mult}}$ & 76.1 & (\g{49.2}) & (\g{70.2})    &  45.2 & 66.2\\
        GNT$\sigma_{0.5}$  & 75.9 & (\g{49.4}) & (\g{70.6})  & 47.1 &   68.3\\
       [0.5em]

       ANT$^{1\mathrm{x}1}$  & 76.0 & (\g{51.1})  & (\g{72.2})   &   47.7   & 68.8 \\   
       ANT$^{1\mathrm{x}1}$+SIN & 74.9 & (\g{52.2}) & (\g{73.6}) & 49.2 & 70.6 \\
       ANT$^{1\mathrm{x}1}$ w/o EP &  75.7 & (\g{48.9}) & (\g{70.2}) & 46.5 & 67.7                   \\ [0.5em]
       
       ANT$^{3\mathrm{x}3}$ & 76.1 & 50.4 & 71.5 & 47.0 & 68.1 \\
       ANT$^{3\mathrm{x}3}$+SIN & 74.1 & \f{52.6} & \f{74.4} & \f{50.6} & \f{72.5} \\
      \end{tabular}
      \caption{Average accuracy on clean data, average Top-1 and Top-5 accuracies on ImageNet-C and ImageNet-C without the Noise categories (higher is better). We compare the results obtained by the means of Gaussian (GNT) and Speckle noise data augmentation and with Adversarial Noise Training (ANT) to several baselines. 
      Gray numbers in brackets indicate scenarios where a corruption from the test set was used during training.}
         \label{tab:IN-C1}
  \end{center}
\end{table}

\subsection{Robustness towards adversarial perturbations}

As regular adversarial training can decrease the accuracy on common corruptions, it is also interesting to check what happens vice-versa:  How does a model which is robust on common corruptions behave under adversarial attacks?

Both our ANT$^{1\mathrm{x}1}$ and GNT models have slightly increased $\ell_2$ and $\ell_{\infty}$ robustness scores compared to a vanilla trained model, see Table~\ref{tab:adv_robustness_in}. We tested this using the white-box attacks PGD \cite{madry2017towards} and DDN \cite{rony2019decoupling}. 
Expectedly, an adversarially trained model has higher adversarial robustness compared to ANT$^{1\mathrm{x}1}$ or GNT. In this experiment, we only verify that we do not unintentionally reduce adversarial robustness compared to a vanilla ResNet50.
For details, see Appendix~E for MNIST and Appendix~F for ImageNet.

    \begin{table}[h!]
      \begin{center}
          \begin{tabular}{l c c c} 
           model & clean acc.\,[\%] &  $\ell_2$ acc.\,[\%] &   $\ell_{\infty}$ acc.\,[\%] \\

            \hline 
             Vanilla RN50  &   75.2  &   41.1  &   18.1 \\
             GNT$\sigma_{0.5}$  &   75.3  &   49.0  &   28.1 \\
             ANT$^{1\mathrm{x}1}$   &   75.7  &   50.1  &   28.6 \\
             Adv. Training \cite{adv_training_free} & 60.5	& \f{58.1} & \f{58.5} \\
      \end{tabular}
      \caption{Adversarial robustness on  $\ell_2$  ($\epsilon = 0.12 $) and $\ell_{\infty}$  ($\epsilon = 0.001 $) compared to a Vanilla ResNet50.}
         \label{tab:adv_robustness_in}
   \end{center}
    \end{table}

\section{Discussion \& Conclusion}

    So far, attempts to use simple noise augmentations for general robustness against common corruptions have  produced mixed results, ranging from no generalization from one noise to other noise types \citep{NIPS2018_7982} to only marginal robustness increases \citep{ford2019adversarial, Lopes_Gaussian_Patch}.
    In this work, we demonstrate that carefully tuned additive noise patterns in conjunction with training on clean samples can surpass almost all current state-of-the-art defense methods against common corruptions. By drawing inspiration from adversarial training and experience replay, we additionally show that training against simple uncorrelated worst-case noise patterns outperforms our already strong baseline defense, with additional gains to be made in combination with previous defense methods like stylization training \citep{geirhos2018imagenettrained}.
    
    There are still a few corruption types (e.g. Motion or Zoom blurs) on which our method is not state of the art, suggesting that additional gains are possible. Future extensions of this work may combine noise generators with varying correlation lengths, add additional interactions between noise and image (e.g.\ multiplicative interactions or local deformations) or take into account local image information in the noise generation process to further boost robustness across many types of image corruptions.

\section{Acknowledgements}
The authors thank the International Max Planck Research School for Intelligent Systems (IMPRS-IS) for supporting Evgenia Rusak and Lukas Schott. The authors thank Yash Sharma for helpful discussions and Alexander Ecker, Robert Geirhos and Dylan Paiton for helpful feed-back while writing the manuscript.
The authors thank the International Max Planck Research School for Intelligent Systems (IMPRS-IS) for supporting Evgenia Rusak, Lukas Schott and Roland S. Zimmermann.
Oliver Bringmann and Evgenia Rusak have been partially supported by the Deutsche Forschungsgemeinschaft (DFG) in the priority program  1835 “Cooperatively Interacting Automobiles” under grant BR2321/5-1 and BR2321/5-2.
This work was supported by the German Federal Ministry of Education and Research (BMBF): Tübingen AI Center, FKZ: 01IS18039A, by the  Deutsche Forschungsgemeinschaft (DFG, German Research Foundation): Germany’s Excellence Strategy – EXC 2064/1 – 390727645 and SFB 1233, Robust Vision: Inference Principles and Neural Mechanisms, TP XX, project number: 276693517 and additionally by the Intelligence Advanced Research Projects Activity (IARPA) via Department of Interior/Interior Business Center (DoI/IBC) contract number D16PC00003. The U.S. Government is authorized to reproduce and distribute reprints for Governmental purposes notwithstanding any copyright annotation thereon. Disclaimer: The views and conclusions contained herein are those of the authors and should not be interpreted as necessarily representing the official policies or endorsements, either expressed or implied, of IARPA, DoI/IBC, or the U.S. Government.
\bibliographystyle{unsrtnat}

\bibliography{references}

\begin{thebibliography}{50}
\providecommand{\natexlab}[1]{#1}
\providecommand{\url}[1]{\texttt{#1}}
\expandafter\ifx\csname urlstyle\endcsname\relax
  \providecommand{\doi}[1]{doi: #1}\else
  \providecommand{\doi}{doi: \begingroup \urlstyle{rm}\Url}\fi

\bibitem[He et~al.(2016)He, Zhang, Ren, and Sun]{he2016deep}
Kaiming He, Xiangyu Zhang, Shaoqing Ren, and Jian Sun.
\newblock Deep residual learning for image recognition.
\newblock In \emph{Proceedings of the IEEE conference on computer vision and
  pattern recognition}, pages 770--778, 2016.

\bibitem[Xiong et~al.(2016)Xiong, Droppo, Huang, Seide, Seltzer, Stolcke, Yu,
  and Zweig]{DBLP:journals/corr/XiongDHSSSYZ16a}
Wayne Xiong, Jasha Droppo, Xuedong Huang, Frank Seide, Mike Seltzer, Andreas
  Stolcke, Dong Yu, and Geoffrey Zweig.
\newblock Achieving human parity in conversational speech recognition.
\newblock \emph{IEEE/ACM Transactions on Audio, Speech, and Language
  Processing}, 2016.

\bibitem[Silver et~al.(2017)Silver, Schrittwieser, Simonyan, Antonoglou, Huang,
  Guez, Hubert, Baker, Lai, Bolton, Chen, Lillicrap, Hui, Sifre, van~den
  Driessche, Graepel, and Hassabis]{Silver2017MasteringTG}
David Silver, Julian Schrittwieser, Karen Simonyan, Ioannis Antonoglou, Aja
  Huang, Arthur Guez, Thomas Hubert, L~Robert Baker, Matthew Lai, Adrian
  Bolton, Yutian Chen, Timothy~P. Lillicrap, Fan Fong~Celine Hui, Laurent
  Sifre, George van~den Driessche, Thore Graepel, and Demis Hassabis.
\newblock Mastering the game of go without human knowledge.
\newblock \emph{Nature}, 550:\penalty0 354--359, 2017.

\bibitem[Campbell et~al.(2002)Campbell, Hoane, and
  Hsu]{Campbell:2002:DB:512148.512152}
Murray Campbell, A.~Joseph Hoane, Jr., and Feng-hsiung Hsu.
\newblock Deep blue.
\newblock \emph{Artif. Intell.}, 134\penalty0 (1-2):\penalty0 57--83, January
  2002.
\newblock ISSN 0004-3702.
\newblock \doi{10.1016/S0004-3702(01)00129-1}.
\newblock URL \url{http://dx.doi.org/10.1016/S0004-3702(01)00129-1}.

\bibitem[OpenAI(2018)]{OpenAI_dota}
OpenAI.
\newblock Openai five.
\newblock \url{https://blog.openai.com/openai-five/}, 2018.

\bibitem[Szegedy et~al.(2013)Szegedy, Zaremba, Sutskever, Bruna, Erhan,
  Goodfellow, and Fergus]{szegedy2013intriguing}
Christian Szegedy, Wojciech Zaremba, Ilya Sutskever, Joan Bruna, Dumitru Erhan,
  Ian Goodfellow, and Rob Fergus.
\newblock Intriguing properties of neural networks.
\newblock \emph{arXiv preprint arXiv:1312.6199}, 2013.

\bibitem[Madry et~al.(2018)Madry, Makelov, Schmidt, Tsipras, and
  Vladu]{madry2018towards}
Aleksander Madry, Aleksandar Makelov, Ludwig Schmidt, Dimitris Tsipras, and
  Adrian Vladu.
\newblock Towards deep learning models resistant to adversarial attacks.
\newblock In \emph{International Conference on Learning Representations}, 2018.
\newblock URL \url{https://openreview.net/forum?id=rJzIBfZAb}.

\bibitem[Schott et~al.(2019)Schott, Rauber, Bethge, and
  Brendel]{schott2018towards}
Lukas Schott, Jonas Rauber, Matthias Bethge, and Wieland Brendel.
\newblock Towards the first adversarially robust neural network model on
  {MNIST}.
\newblock In \emph{International Conference on Learning Representations}, 2019.
\newblock URL \url{https://openreview.net/forum?id=S1EHOsC9tX}.

\bibitem[Gilmer et~al.(2018)Gilmer, Metz, Faghri, Schoenholz, Raghu,
  Wattenberg, and Goodfellow]{DBLP:journals/corr/abs-1801-02774}
Justin Gilmer, Luke Metz, Fartash Faghri, Samuel~S. Schoenholz, Maithra Raghu,
  Martin Wattenberg, and Ian~J. Goodfellow.
\newblock Adversarial spheres.
\newblock \emph{CoRR}, abs/1801.02774, 2018.
\newblock URL \url{http://arxiv.org/abs/1801.02774}.

\bibitem[Geirhos et~al.(2018)Geirhos, Temme, Rauber, Sch\"{u}tt, Bethge, and
  Wichmann]{NIPS2018_7982}
Robert Geirhos, Carlos R.~M. Temme, Jonas Rauber, Heiko~H. Sch\"{u}tt, Matthias
  Bethge, and Felix~A. Wichmann.
\newblock Generalisation in humans and deep neural networks.
\newblock In S.~Bengio, H.~Wallach, H.~Larochelle, K.~Grauman, N.~Cesa-Bianchi,
  and R.~Garnett, editors, \emph{Advances in Neural Information Processing
  Systems 31}, pages 7538--7550. Curran Associates, Inc., 2018.
\newblock URL
  \url{http://papers.nips.cc/paper/7982-generalisation-in-humans-and-deep-neural-networks.pdf}.

\bibitem[Ford et~al.(2019)Ford, Gilmer, Carlini, and
  Cubuk]{ford2019adversarial}
Nic Ford, Justin Gilmer, Nicolas Carlini, and Dogus Cubuk.
\newblock Adversarial examples are a natural consequence of test error in
  noise.
\newblock \emph{ICML}, 2019.

\bibitem[Russakovsky et~al.(2014)Russakovsky, Deng, Su, Krause, Satheesh, Ma,
  Huang, Karpathy, Khosla, Bernstein, Berg, and
  Li]{DBLP:journals/corr/RussakovskyDSKSMHKKBBF14}
Olga Russakovsky, Jia Deng, Hao Su, Jonathan Krause, Sanjeev Satheesh, Sean Ma,
  Zhiheng Huang, Andrej Karpathy, Aditya Khosla, Michael~S. Bernstein,
  Alexander~C. Berg, and Fei{-}Fei Li.
\newblock Imagenet large scale visual recognition challenge.
\newblock \emph{CoRR}, abs/1409.0575, 2014.
\newblock URL \url{http://arxiv.org/abs/1409.0575}.

\bibitem[Hendrycks and Dietterich(2019)]{hendrycks2018benchmarking}
Dan Hendrycks and Thomas Dietterich.
\newblock Benchmarking neural network robustness to common corruptions and
  perturbations.
\newblock In \emph{International Conference on Learning Representations}, 2019.
\newblock URL \url{https://openreview.net/forum?id=HJz6tiCqYm}.

\bibitem[Michaelis et~al.(2019)Michaelis, Mitzkus, Geirhos, Rusak, Bringmann,
  Ecker, Bethge, and Brendel]{michaelis2019benchmarking}
Claudio Michaelis, Benjamin Mitzkus, Robert Geirhos, Evgenia Rusak, Oliver
  Bringmann, Alexander~S Ecker, Matthias Bethge, and Wieland Brendel.
\newblock Benchmarking robustness in object detection: Autonomous driving when
  winter is coming.
\newblock \emph{arXiv preprint arXiv:1907.07484}, 2019.

\bibitem[Mu and Gilmer(2019)]{mu2019mnist}
Norman Mu and Justin Gilmer.
\newblock \uppercase{MNIST-C}: A robustness benchmark for computer vision.
\newblock \emph{arXiv preprint arXiv:1906.02337}, 2019.

\bibitem[Dodge and Karam(2016)]{DBLP:journals/corr/DodgeK16}
Samuel~Fuller Dodge and Lina~J. Karam.
\newblock Understanding how image quality affects deep neural networks.
\newblock \emph{2016 Eighth International Conference on Quality of Multimedia
  Experience (QoMEX)}, pages 1--6, 2016.

\bibitem[Dodge and Karam(2017{\natexlab{a}})]{DBLP:journals/corr/DodgeK17b}
Samuel~F. Dodge and Lina~J. Karam.
\newblock A study and comparison of human and deep learning recognition
  performance under visual distortions.
\newblock \emph{CoRR}, abs/1705.02498, 2017{\natexlab{a}}.
\newblock URL \url{http://arxiv.org/abs/1705.02498}.

\bibitem[Yin et~al.(2020)Yin, Lopes, Shlens, Cubuk, and
  Gilmer]{DBLP:journals/corr/abs-1906-08988}
Dong Yin, Raphael~Gontijo Lopes, Jonathon Shlens, Ekin~D. Cubuk, and Justin
  Gilmer.
\newblock A \uppercase{F}ourier perspective on model robustness in computer
  vision.
\newblock \emph{NeurIPS}, 2020.

\bibitem[Zhang(2019)]{zhang2019making}
Richard Zhang.
\newblock Making convolutional networks shift-invariant again.
\newblock \emph{ICML}, 2019.

\bibitem[Xie et~al.(2019{\natexlab{a}})Xie, Hovy, Luong, and
  Le]{xie2019selftraining}
Qizhe Xie, Eduard Hovy, Minh-Thang Luong, and Quoc~V. Le.
\newblock Self-training with noisy student improves imagenet classification.
\newblock \emph{arXiv preprint arXiv:1911.04252}, 2019{\natexlab{a}}.

\bibitem[Mahajan et~al.(2018)Mahajan, Girshick, Ramanathan, He, Paluri, Li,
  Bharambe, and van~der Maaten]{mahajan2018exploring}
Dhruv Mahajan, Ross Girshick, Vignesh Ramanathan, Kaiming He, Manohar Paluri,
  Yixuan Li, Ashwin Bharambe, and Laurens van~der Maaten.
\newblock Exploring the limits of weakly supervised pretraining.
\newblock In \emph{Proceedings of the European Conference on Computer Vision
  (ECCV)}, pages 181--196, 2018.

\bibitem[Mikołajczyk and Grochowski(2018)]{Mikoajczyk2018DataAF}
Agnieszka Mikołajczyk and Michał Grochowski.
\newblock Data augmentation for improving deep learning in image classification
  problem.
\newblock \emph{2018 International Interdisciplinary PhD Workshop (IIPhDW)},
  pages 117--122, 2018.

\bibitem[Geirhos et~al.(2019)Geirhos, Rubisch, Michaelis, Bethge, Wichmann, and
  Brendel]{geirhos2018imagenettrained}
Robert Geirhos, Patricia Rubisch, Claudio Michaelis, Matthias Bethge, Felix~A.
  Wichmann, and Wieland Brendel.
\newblock Imagenet-trained {CNN}s are biased towards texture; increasing shape
  bias improves accuracy and robustness.
\newblock In \emph{International Conference on Learning Representations}, 2019.
\newblock URL \url{https://openreview.net/forum?id=Bygh9j09KX}.

\bibitem[Dodge and Karam(2017{\natexlab{b}})]{DBLP:journals/corr/DodgeK17a}
Samuel~F. Dodge and Lina~J. Karam.
\newblock Quality resilient deep neural networks.
\newblock \emph{CoRR}, abs/1703.08119, 2017{\natexlab{b}}.
\newblock URL \url{http://arxiv.org/abs/1703.08119}.

\bibitem[Hendrycks et~al.(2020)Hendrycks, Mu, Cubuk, Zoph, Gilmer, and
  Lakshminarayanan]{hendrycks2020augmix}
Dan Hendrycks, Norman Mu, Ekin~Dogus Cubuk, Barret Zoph, Justin Gilmer, and
  Balaji Lakshminarayanan.
\newblock Augmix: A simple data processing method to improve robustness and
  uncertainty.
\newblock In \emph{International Conference on Learning Representations}, 2020.
\newblock URL \url{https://openreview.net/forum?id=S1gmrxHFvB}.

\bibitem[Cohen et~al.(2019)Cohen, Rosenfeld, and Kolter]{pmlr-v97-cohen19c}
Jeremy Cohen, Elan Rosenfeld, and Zico Kolter.
\newblock Certified adversarial robustness via randomized smoothing.
\newblock In Kamalika Chaudhuri and Ruslan Salakhutdinov, editors,
  \emph{Proceedings of the 36th International Conference on Machine Learning},
  volume~97 of \emph{Proceedings of Machine Learning Research}, pages
  1310--1320, Long Beach, California, USA, 09--15 Jun 2019. PMLR.

\bibitem[Lopes et~al.(2019)Lopes, Yin, Poole, Gilmer, and
  Cubuk]{Lopes_Gaussian_Patch}
Raphael~Gontijo Lopes, Dong Yin, Ben Poole, Justin Gilmer, and Ekin~D. Cubuk.
\newblock Improving robustness without sacrificing accuracy with patch gaussian
  augmentation.
\newblock \emph{CoRR}, abs/1906.02611, 2019.
\newblock URL \url{http://arxiv.org/abs/1906.02611}.

\bibitem[Laermann et~al.(2019)Laermann, Samek, and
  Strodthoff]{laermann2019achieving}
Jan Laermann, Wojciech Samek, and Nils Strodthoff.
\newblock Achieving generalizable robustness of deep neural networks by
  stability training.
\newblock In \emph{German Conference on Pattern Recognition}, pages 360--373.
  Springer, 2019.

\bibitem[Fawzi et~al.(2016)Fawzi, Moosavi{-}Dezfooli, and
  Frossard]{DBLP:journals/corr/FawziMF16}
Alhussein Fawzi, Seyed{-}Mohsen Moosavi{-}Dezfooli, and Pascal Frossard.
\newblock Robustness of classifiers: from adversarial to random noise.
\newblock \emph{NIPS}, 2016.

\bibitem[Engstrom et~al.(2019)Engstrom, Tsipras, Schmidt, and
  Madry]{engstrom2017rotation}
Logan Engstrom, Dimitris Tsipras, Ludwig Schmidt, and Aleksander Madry.
\newblock A rotation and a translation suffice: Fooling cnns with simple
  transformations.
\newblock \emph{ICML}, 2019.

\bibitem[Kang et~al.(2019)Kang, Sun, Brown, Hendrycks, and
  Steinhardt]{kang2019transfer}
Daniel Kang, Yi~Sun, Tom Brown, Dan Hendrycks, and Jacob Steinhardt.
\newblock Transfer of adversarial robustness between perturbation types.
\newblock \emph{CoRR}, abs/1905.01034, 2019.
\newblock URL \url{http://arxiv.org/abs/1905.01034}.

\bibitem[Xie et~al.(2019{\natexlab{b}})Xie, Wu, van~der Maaten, Yuille, and
  He]{featdenoise}
Cihang Xie, Yuxin Wu, Laurens van~der Maaten, Alan~L. Yuille, and Kaiming He.
\newblock Feature denoising for improving adversarial robustness.
\newblock \emph{CVPR}, 2019{\natexlab{b}}.

\bibitem[Jordan et~al.(2019)Jordan, Manoj, Goel, and
  Dimakis]{jordan2019quantifying}
Matt Jordan, Naren Manoj, Surbhi Goel, and Alexandros~G Dimakis.
\newblock Quantifying perceptual distortion of adversarial examples.
\newblock \emph{arXiv preprint arXiv:1902.08265}, 2019.

\bibitem[Tram{\`{e}}r and Boneh(2019)]{Tramer2019Adversarial}
Florian Tram{\`{e}}r and Dan Boneh.
\newblock Adversarial training and robustness for multiple perturbations.
\newblock \emph{NeurIPS}, 2019.
\newblock URL \url{http://arxiv.org/abs/1904.13000}.

\bibitem[Moosavi{-}Dezfooli et~al.(2017)Moosavi{-}Dezfooli, Fawzi, Fawzi, and
  Frossard]{DBLP:journals/corr/Moosavi-Dezfooli16}
Seyed{-}Mohsen Moosavi{-}Dezfooli, Alhussein Fawzi, Omar Fawzi, and Pascal
  Frossard.
\newblock Universal adversarial perturbations.
\newblock \emph{CVPR}, 2017.

\bibitem[Hayes and Danezis(2017)]{DBLP:journals/corr/abs-1708-05207}
Jamie Hayes and George Danezis.
\newblock Machine learning as an adversarial service: Learning black-box
  adversarial examples.
\newblock \emph{CoRR}, abs/1708.05207, 2017.
\newblock URL \url{http://arxiv.org/abs/1708.05207}.

\bibitem[Metzen(2018)]{hendrik2018universality}
Jan~Hendrik Metzen.
\newblock Universality, robustness, and detectability of adversarial
  perturbations under adversarial training.
\newblock \emph{https://openreview.net/forum?id=SyjsLqxR-}, 2018.

\bibitem[Shafahi et~al.(2018)Shafahi, Najibi, Xu, Dickerson, Davis, and
  Goldstein]{DBLP:journals/corr/abs-1811-11304}
Ali Shafahi, Mahyar Najibi, Zheng Xu, John~P. Dickerson, Larry~S. Davis, and
  Tom Goldstein.
\newblock Universal adversarial training.
\newblock \emph{CoRR}, abs/1811.11304, 2018.
\newblock URL \url{http://arxiv.org/abs/1811.11304}.

\bibitem[Mummadi et~al.(2019)Mummadi, Brox, and
  Metzen]{DBLP:journals/corr/abs-1812-03705}
Chaithanya~Kumar Mummadi, Thomas Brox, and Jan~Hendrik Metzen.
\newblock Defending against universal perturbations with shared adversarial
  training.
\newblock \emph{ICCV}, 2019.

\bibitem[P{\'{e}}rolat et~al.(2018)P{\'{e}}rolat, Malinowski, Piot, and
  Pietquin]{DBLP:journals/corr/abs-1809-07802}
Julien P{\'{e}}rolat, Mateusz Malinowski, Bilal Piot, and Olivier Pietquin.
\newblock Playing the game of universal adversarial perturbations.
\newblock \emph{CoRR}, abs/1809.07802, 2018.
\newblock URL \url{http://arxiv.org/abs/1809.07802}.

\bibitem[Rauber and Bethge(2020)]{rauber2020fast}
Jonas Rauber and Matthias Bethge.
\newblock Fast differentiable clipping-aware normalization and rescaling.
\newblock \emph{arXiv preprint arXiv:2007.07677}, 2020.
\newblock URL \url{https://github.com/jonasrauber/clipping-aware-rescaling}.

\bibitem[Mnih et~al.(2015)Mnih, Kavukcuoglu, Silver, Rusu, Veness, Bellemare,
  Graves, Riedmiller, Fidjeland, Ostrovski, et~al.]{mnih2015human}
Volodymyr Mnih, Koray Kavukcuoglu, David Silver, Andrei~A Rusu, Joel Veness,
  Marc~G Bellemare, Alex Graves, Martin Riedmiller, Andreas~K Fidjeland, Georg
  Ostrovski, et~al.
\newblock Human-level control through deep reinforcement learning.
\newblock \emph{Nature}, 518\penalty0 (7540):\penalty0 529, 2015.

\bibitem[Paszke et~al.(2017)Paszke, Gross, Chintala, Chanan, Yang, DeVito, Lin,
  Desmaison, Antiga, and Lerer]{paszke2017automatic}
Adam Paszke, Sam Gross, Soumith Chintala, Gregory Chanan, Edward Yang, Zachary
  DeVito, Zeming Lin, Alban Desmaison, Luca Antiga, and Adam Lerer.
\newblock Automatic differentiation in {PyTorch}.
\newblock In \emph{NIPS Autodiff Workshop}, 2017.

\bibitem[Engstrom et~al.(2018)Engstrom, Ilyas, and Athalye]{logitpairing2018}
Logan Engstrom, Andrew Ilyas, and Anish Athalye.
\newblock Evaluating and understanding the robustness of adversarial logit
  pairing.
\newblock \emph{CoRR}, abs/1807.10272, 2018.
\newblock URL \url{https://arxiv.org/abs/1807.10272}.

\bibitem[Shafahi et~al.(2019)Shafahi, Najibi, Ghiasi, Xu, Dickerson, Studer,
  Davis, Taylor, and Goldstein]{adv_training_free}
Ali Shafahi, Mahyar Najibi, Amin Ghiasi, Zheng Xu, John Dickerson, Christoph
  Studer, Larry~S Davis, Gavin Taylor, and Tom Goldstein.
\newblock Adversarial training for free!
\newblock \emph{arXiv preprint arXiv:1904.12843}, 2019.

\bibitem[Cubuk et~al.(2018)Cubuk, Zoph, Mane, Vasudevan, and
  Le]{cubuk2018autoaugment}
Ekin~D Cubuk, Barret Zoph, Dandelion Mane, Vijay Vasudevan, and Quoc~V Le.
\newblock Autoaugment: Learning augmentation policies from data.
\newblock \emph{arXiv preprint arXiv:1805.09501}, 2018.

\bibitem[Madry et~al.(2017)Madry, Makelov, Schmidt, Tsipras, and
  Vladu]{madry2017towards}
Aleksander Madry, Aleksandar Makelov, Ludwig Schmidt, Dimitris Tsipras, and
  Adrian Vladu.
\newblock Towards deep learning models resistant to adversarial attacks.
\newblock \emph{arXiv preprint arXiv:1706.06083}, 2017.

\bibitem[Rony et~al.(2019)Rony, Hafemann, Oliveira, Ayed, Sabourin, and
  Granger]{rony2019decoupling}
J{\'e}r{\^o}me Rony, Luiz~G Hafemann, Luiz~S Oliveira, Ismail~Ben Ayed, Robert
  Sabourin, and Eric Granger.
\newblock Decoupling direction and norm for efficient gradient-based l2
  adversarial attacks and defenses.
\newblock In \emph{Proceedings of the IEEE Conference on Computer Vision and
  Pattern Recognition}, pages 4322--4330, 2019.

\bibitem[Kingma and Ba(2014)]{kingma2014adam}
Diederik~P Kingma and Jimmy Ba.
\newblock Adam: A method for stochastic optimization.
\newblock \emph{arXiv preprint arXiv:1412.6980}, 2014.

\bibitem[Wong et~al.(2018)Wong, Schmidt, Metzen, and
  Kolter]{DBLP:journals/corr/abs-1805-12514}
Eric Wong, Frank~R. Schmidt, Jan~Hendrik Metzen, and J.~Zico Kolter.
\newblock Scaling provable adversarial defenses.
\newblock \emph{CoRR}, abs/1805.12514, 2018.
\newblock URL \url{http://arxiv.org/abs/1805.12514}.

\end{thebibliography}

\newpage
\renewcommand{\thesubsection}{\Alph{subsection}}
\section*{Appendix}
\appendix
\subsection{Architectures of the noise generators}

The architectures of the noise generators are displayed in Tables~\ref{tab:arch1} \ref{tab:arch1} and \ref{tab:arch2}. The number of color channels is indicated by $C$. The noise generator displayed in Table~\ref{tab:arch1} only uses kernels with a size of 1 and thus produces spatially uncorrelated noise. With the stride being 1 and no padding, the spatial dimensions are preserved in each layer. The noise generator displayed in Table~\ref{tab:arch2} has one layer with 3x3 convolutions and thus produces noise samples with a correlation length of 3x3 pixels.

\begin{table}[h]
\begin{center}
\parbox{.4\linewidth}{
    \centering
        \begin{tabular}{lc}
            Layer & Shape \\
            \hline\\[-1em] 
            Conv + ReLU & $20\times1\times1$ \\
            Conv + ReLU & $20\times1\times1$ \\
            Conv + ReLU & $20\times1\times1$ \\
            Conv & C $\times1\times1$ \\
        \end{tabular}
   \caption{Architecture of the noise generator producing uncorrelated noise.}
  \label{tab:arch1}
}
\parbox{.4\linewidth}{
        \centering
            \begin{tabular}{lc}
                Layer & Shape \\
                \hline\\[-1em]
                Conv + ReLU & $20\times1\times1$ \\
                Conv + ReLU & $20\times3\times3$ \\
                Conv + ReLU & $20\times1\times1$ \\
                Conv & C $\times1\times1$ \\
            \end{tabular}
      \caption{Architecture of the noise generator producing locally correlated noise.}
      \label{tab:arch2}
}
\end{center}

\end{table}

\subsection{Implementation details and hyper-parameters}

\paragraph{Preprocessing}
MNIST images are preprocessed such that their pixel values lie in the range $[0, 1]$.
Preprocessing for ImageNet is performed in the standard way for PyTorch ImageNet models from the model zoo by subtracting the mean $[0.485, 0.456, 0.406]$ and dividing by the standard deviation $[0.229, 0.224, 0.225]$.
We add Gaussian, adversarial and Speckle noise before the preprocessing step, so the noisy images are first clipped to the range $[0,1]$ of the raw images and then preprocessed before being fed into the model.

\subsubsection{ImageNet experiments}
For all ImageNet experiments, we used a pretrained ResNet50 architecture from \url{https://pytorch.org/docs/stable/torchvision/models.html}. We fine-tuned the model with SGD-M using an initial learning rate of $0.001$, which corresponds to the last learning rate of the PyTorch model training, and a momentum of $0.9$. After convergence, we decayed the learning rate once by a factor of $10$ and continued the training. Decaying the learning rate was highly beneficial for the model performance. We tried decaying the learning rate a second time, but this did not bring any benefits in any of our experiments. 
For GNT, we also tried training from scratch, i.e. starting with a large learning rate of 0.1 and random weights, and trained for 120 epochs, but we got worse results compared to merely fine-tuning the model provided by torchvision.
We used a batch size of $70$ for all our experiments. We have also tried to use the batch sizes $50$ and $100$, but did not see major effects.

\paragraph{Gaussian noise}
We trained the models until convergence. The total number of training epochs varied between $30$ and $90$ epochs.

\paragraph{Speckle noise}
We used the Speckle noise implementation from \url{https://github.com/hendrycks/robustness/blob/master/ImageNet-C/create_c/make_imagenet_c.py}, line $270$. The model trained with Speckle noise converged faster than with Gaussian data augmentation and therefore, we only trained the model for $10$ epochs.

\paragraph{Adversarial Noise Training}
The adversarial noise generator was trained with the Adam optimizer with a learning rate of $0.0001$. We have replaced the noise generator every $0.33$ epochs. For ANT$^{1\mathrm{x}1}$, we set the $\epsilon$-sphere to control the size of the perturbation to $135.0$ which on average corresponds to the $\ell_2$-size of a perturbation caused by additive Gaussian noise sampled from $\mathcal{N}(0, 0.5^2 \cdot \mathds{1})$. We have trained the classifier until convergence for $80$ epochs. For ANT$^{3\mathrm{x}3}$, we set the $\epsilon$-sphere to $70.0$ and trained the classifier for 80 epochs. We decreased the $\epsilon$-sphere for ANT$^{3\mathrm{x}3}$ to counteract giving the noise generator more degrees of freedom to fool the classifier to maintain a similar training losses and accuracies for ANT$^{1\mathrm{x}1}$ and ANT$^{3\mathrm{x}3}$.

\subsubsection{MNIST experiments}
For the MNIST experiments, we used the same model architecture as \citet{madry2017towards} for our ANT$^{1\mathrm{x}1}$ and GNT.
For ANT$^{1\mathrm{x}1}$, our learning rate for the generator was between $10^{-4}$ and $10^{-5}$, and equal to $10^{-3}$ for the classifier. We used a batch size of $300$. As an optimizer, we used SGD-M with a momentum of $0.9$ for the classifier and Adam \citep{kingma2014adam} for the generator. The splitting of batches in clean, noisy and history was equivalent to the ImageNet experiments. The optimal $\epsilon$ hyper-parameter was determined with a line search similar to the optimal $\sigma$ of the Gaussian noise; we found $\epsilon = 10$ to be optimal. The parameters for the Gaussian noise experiments were equivalent.  Both models were trained until convergence (around 500-600 epochs). GNT and ANT$^{1\mathrm{x}1}$ were performed on a pretrained network.

\subsection{Detailed results on the evaluation of robustness due to regular adversarial training}

We find that standard adversarial training against minimal adversarial perturbations in general does not increase robustness against common corruptions. While some early results on CIFAR-10 by \citet{ford2019adversarial} and Tiny ImageNet-C by \citet{hendrycks2018benchmarking} suggest that standard adversarial training might increase robustness to common corruptions, we here observe the opposite:
Adversarially trained models have lower robustness against common corruptions.
An adversarially trained ResNet152 with an additional denoising layer\footnote{Model weights from \mbox{\url{https://github.com/facebookresearch/ImageNet-Adversarial-Training}}} from \citet{featdenoise} has lower accuracy across almost all corruptions except Snow and Pixelations. On some corruptions, the accuracy of the adversarially trained model decreases drastically, e.g.\ from $49.1\%$ to $4.6\%$ on Fog or $42.8\%$ to $9.3\%$ on Contrast.
Similarly, the adversarially trained ResNet50\footnote{\label{note1}Model weights were kindly provided by the authors.} from [Shafahi et al., 2019] shows a substantial decrease in performance on common corruptions compared with a vanilla trained model.\\

An evaluation of a robustified version of AlexNet\footnoteref{note1} \citep{DBLP:journals/corr/abs-1811-11304} that was trained with the Universal Adversarial Training scheme on ImageNet-C shows that achieving robustness against universal adversarial perturbations does not noticeably increase robustness towards common corruptions ($22.2\%$) compared with a vanilla trained model ($21.1\%$).

\begin{table}[ht]\footnotesize
  \begin{center}
      \begin{tabular}{l|c|c c c|c c c c} 
        &  &  \multicolumn{3}{c|}{Noise (Compressed)}  &  \multicolumn{4}{c}{Blur (Compressed)} \\
        Model              & All & Gaussian & Shot& Impulse& Defocus& Glass   & Motion &  Zoom \\
        \hline 
        Vanilla RN50                   & 39.2     & 29.3     & 27.0     & 23.8     & 38.7     & 26.8     & 38.7     & 36.2   \\
        AT \citep{adv_training_free}  & 29.1 & 20.5 & 19.1 & 12.4 & 21.4 & 30.8 & 30.4 & 31.4 \\
       \hline
        Vanilla RN152                  & 45.0     & 35.7     & 34.3     & 29.6     & 45.1     & 32.8     & 48.4     & 40.5 \\
        AT \citep{featdenoise}                 & 35.0     & 35.2     & 34.4     & 24.8     & 22.1     & 31.7     & 30.9     & 32.0    \\  
         \hline
        Vanilla AlexNet      & 21.1 & 11.4 & 10.6 & 7.7 & 18.0 & 17.4 & 21.4 & 20.2 \\
        UAT \citep{DBLP:journals/corr/abs-1811-11304} & 22.2 & 20.1 & 19.1 & 16.2 & 13.1 & 21.6 & 19.7 & 19.2 \\
      \end{tabular}
\newline
\vspace*{0.5 cm}
\newline
      \begin{tabular}{l|c c c c|c c c c} 
        &   \multicolumn{4}{c|}{Weather (Compressed)}  &  \multicolumn{4}{c}{Digital (Compressed)} \\
        Model               & Snow   & Frost & Fog      &Brightness & Contrast & Elastic & Pixelate &  JPEG \\
        \hline 
        Vanilla RN50                    & 32.5     & 38.1     & 45.8     & 68.0     & 39.1     & 45.2     & 44.8     & 53.4  \\
        AT \citep{adv_training_free} & 24.4 & 25.6 & 5.8 & 51.1 & 7.8 & 45.4 & 53.4 & 56.3 \\
       \hline
        Vanilla RN152                   & 38.7     & 43.9     & 49.1     & 71.2     & 42.8     & 51.1     & 50.5     & 60.5 \\
       AT \citep{featdenoise}                  & 42.0     & 40.4     & 4.6      & 58.8     & 9.3      & 47.2     & 54.1     & 58.0   \\
       \hline
       Vanilla AlexNet & 13.3 & 17.3 & 18.1 & 43.5 & 14.7 & 35.4 & 28.2 & 39.4 \\
       UAT \citep{DBLP:journals/corr/abs-1811-11304} & 13.8 & 18.3 & 4.3 & 36.5 & 4.8 & 36.8 & 42.3 & 47.1 \\
      \end{tabular}
      \caption{Average Top-1 accuracy over 5 severities of common corruptions on ImageNet-C in percent. A high accuracy on a certain corruption type indicates high robustness of a classifier on this corruption type, so higher accuracy is better. Adversarial training (AT) decreases the accuracy on common corruptions, especially on the corruptions Fog and Contrast. Universal Adversarial Training (UAT) slightly increases the overall performance.}
         \label{tab:IN-C}
  \end{center}
\end{table}

\clearpage
\subsection{Detailed ImageNet-C results}
\begin{figure*}
    \begin{center}
      \includegraphics[width=0.95\textwidth]{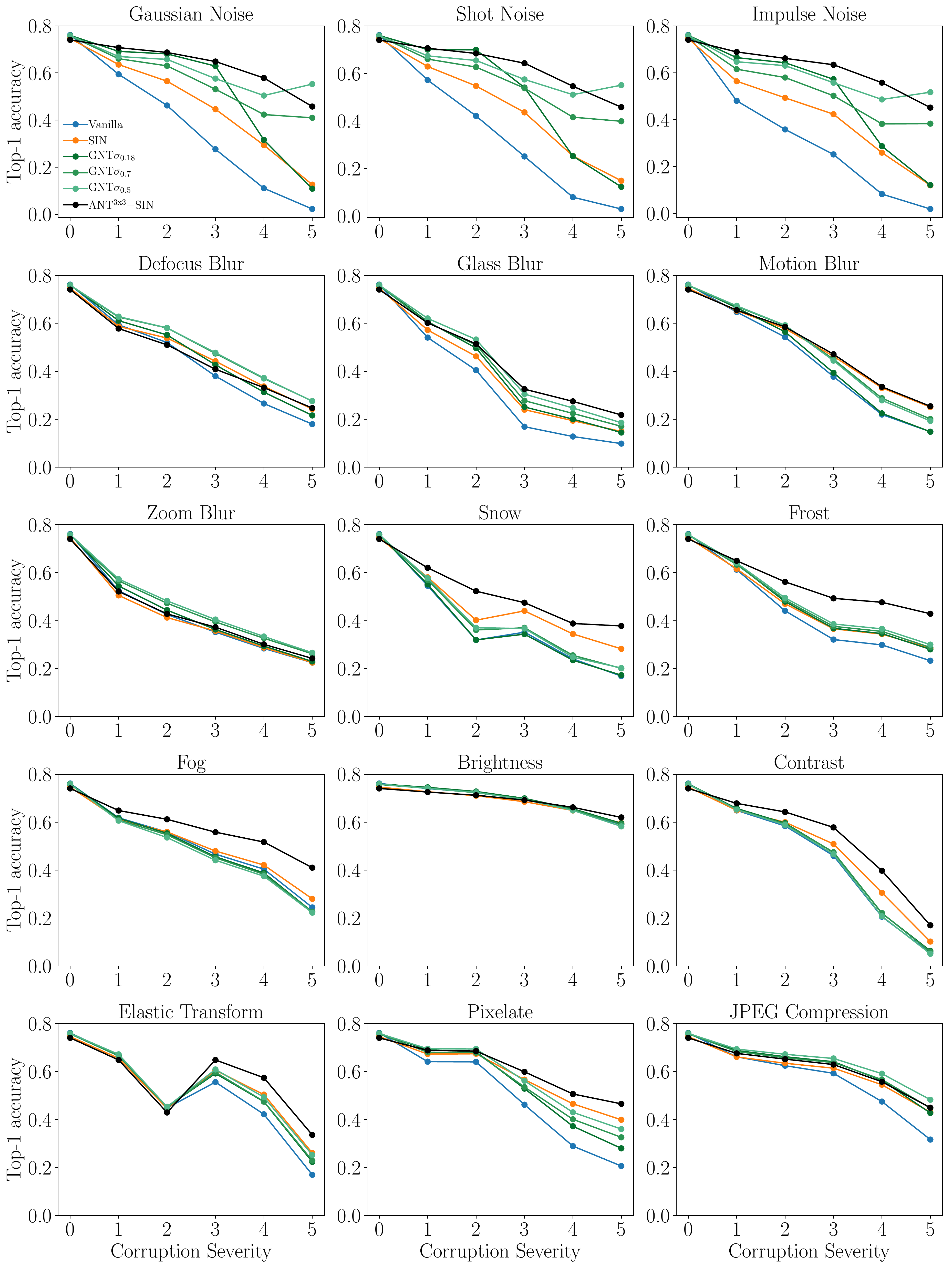}
      \caption{Top-1 accuracy for each corruption type and severity on ImageNet-C.}
      \label{Fig:severities}
      \end{center}
    \end{figure*}

We show detailed results on individual corruptions in Table~\ref{tab:IN-C2} in accuracy and in Table~\ref{tab:IN-C3} in mCE for differently trained models. In Fig.~\ref{Fig:severities}, we show the degradation of accuracy for different severity levels. To avoid clutter, we only show results for a vanilla trained model, for the previous state of the art SIN+IN \citep{geirhos2018imagenettrained}, for several Gaussian trained models and for the overall best model ANT$^{3\mathrm{x}3}$+SIN.

The Corruption Error \citep{hendrycks2018benchmarking} is defined as
\begin{equation}
    \mathrm{CE}_c^f = \left( \sum_{s=1}^5 E_{s,c}^f \right) \bigg/  \left( \sum_{s=1}^5 E_{s,c}^\mathrm{AlexNet} \right),
\end{equation}
where $E_{s,c}^f$ is the Top-1 error of a classifier $f$ for a corruption $c$ with severity $s$. The mean Corruption error (mCE) is taken by averaging over all corruptions.

    \begin{table*}[h]\footnotesize
      \begin{center}
      \setlength\tabcolsep{0.5pt} 
          \begin{tabular}{l c |c c c|c c c c|c c c c|c c c c} 
            & &    \multicolumn{3}{c|}{Noise}  &  \multicolumn{4}{c}{Blur} &   \multicolumn{4}{|c|}{Weather}  &  \multicolumn{4}{c}{Digital} \\
            model      & mean         & Gauss & Shot& Impulse& Defocus& Glass   & Motion &  Zoom & Snow   & Frost & Fog      &Bright & Contrast & Elastic & Pixel &  Jpeg \\
            \hline 
            Vanilla RN50 & 39  & 29     & 27     & 24 & 39 & 27  & 39 & 36 &  33 & 38    & 46 & 68   & 39     & 45     & 45    & 53  \\
            Shift Inv & 42  &  36 & 34 & 30 & 40 & 29 & 38 & 39 & 33 & 40 & 48 & 68 & 42 & 45 & 49 & 57 \\
            Patch GN & 44  &  45 & 43 & 42 & 38 & 26 & 39 & 38 & 30 & 39 & 54 & 67 & 39 & 52 & 47 & 56 \\
            SIN+IN  & 45       & 41     & 40     & 37     & 43  &32  & 45  & 36 & 41  & 42     &47 &67     &43  &50  &56  &58\\
            AugMix      &  48 & 41 &  41 & 38 & \f{48} & 35 & \f{54} & \f{49} & 40 & 44 & 47 & 69 & \f{51} & 52 & 57 & 60 \\[0.5em]

             Speckle &    46 &  55 & 58 & 49 & 43 & 32 & 40 & 36 & 34 & 41 & 46 & 68 & 41 & 47 & 49 & 58 \\ 
            GNT$_{\mathrm{mult}}$ & 49  &  \f{67} & 65 & \f{64} & 43 & 33 & 41 & 37 & 34 & 42 & 45 & 68 & 41 & 48 & 50 & 60 \\
            GNT$\sigma_{0.5}$  & 49  &  58 & 59 & 57 & 47 & 38 & 43 & 42 & 35 & 44 & 44 & 68 & 39 & 50 & 55 & \f{62} \\
         GNT$\sigma_{0.5}$+SIN & 52 &  57 & 58 & 54 & 41 & 39 & 47 & 36 & 48 & 49 & 57 & 67 & 53 & 54 & 57 & 58 \\[0.5em]

            ANT$^{1\mathrm{x}1}$ & 51 & 65 & \f{66} & \f{64} & 47 & 37 & 43 & 40 & 36 & 46 & 44 & \f{70} & 43 & 49 & 55 & \f{62} \\ 

            ANT$^{1\mathrm{x}1}$+SIN & 52 & 64 & 65 & 63 & 46 & 38 & 46 & 39 & 42 & 47 & 49 & 69 & 47 & 50 & \f{57} & 60 \\
            ANT$^{1\mathrm{x}1}$ w/o EP & 49 &  59 & 59 & 57 & 46 & 37 & 43 & 40 & 34 & 43 & 43 & 68 & 39 & 49 & 55 & 61 \\[0.5em]
            
             ANT$^{3\mathrm{x}3}$ & 50 &  65 & 64 & \f{64} & 44 & 36 & 42 & 38 & 39 & 46 & 44 & 69 & 41 & 49 & 55 & 61  \\
            ANT$^{3\mathrm{x}3}$+SIN & \f{53} &  62 & 61 & 60 & 41 & \f{39} & 46 & 37 & \f{48} & \f{52} & \f{55} & 68 & 49 & \f{53} & \f{59} & 59  \\
          \end{tabular}
          \caption{Average Top-1 accuracy over 5 severities of common corruptions on ImageNet-C in percent obtained by different models; higher is better. 
          }

             \label{tab:IN-C2}
      \end{center}
    \end{table*}

\begin{table}[ht]\footnotesize
  \begin{center}
  \setlength\tabcolsep{0.5pt} 
      \begin{tabular}{l c|c c c|c c c c|c c c c|c c c c} 
        &   &  \multicolumn{3}{c|}{Noise}  &  \multicolumn{4}{c}{Blur} &   \multicolumn{4}{|c|}{Weather}  &  \multicolumn{4}{c}{Digital} \\
        model      & mCE        & Gauss & Shot& Impulse& Defocus& Glass   & Motion &  Zoom & Snow   & Frost & Fog      &Bright & Contrast & Elastic & Pixel &  Jpeg \\
        \hline 
        Vanilla & 77 & 80 & 82 & 83 & 75 & 89 & 78 & 80 & 78 & 75 & 66 & 57 & 71 & 85 & 77 & 77\\
        SIN & 69 & 66 & 67 & 68 & 70 & 82 & 69 & 80 & 68 & 71 & 65 & 58 & 66 & 78 & 62 & 70 \\
        Patch GN & 71 & 62 & 63 & 62 & 75 & 90 & 78 & 78 & 81 & 74 & 57 & 59 & 71 & 74 & 74 & 72\\
        Shift Inv. & 73 & 73 & 74 & 76 & 74 & 86 & 78 & 77 & 77 & 72 & 63 & 56 & 68 & 86 & 71 & 71 \\
        AugMix & 65 & 67 & 66 & 68 & \f{64} & 79 & \f{59} & \f{64} & 69 & 68 & 65 & 54 & \f{57} & 74 & 60 & 65 \\[0.5em]

        Speckle & 68 & 51 & 47 & 55 & 70 & 83 & 77 & 80 & 76 & 71 & 66 & 57 & 70 & 82 & 71 & 69 \\
        GNT$_{\mathrm{mult}}$ & 65 & \f{37} & 39 & \f{39} & 69 & 81 & 76 & 79 & 76 & 70 & 67 & 56 & 69 & 81 & 69 & 66 \\
        GNT$\sigma_{0.5}$  & 64 & 46 & 46 & 47 & 65 & 75 & 72 & 74 & 75 & 68 & 69 & 57 & 71 & 78 & 63 & 63 \\
[0.5em]
        
       ANT$^{1\mathrm{x}1}$ & 62 & 39 & \f{38} & \f{39} & 65 & 77 & 72 & 75 & 74 & 66 & 68 & \f{53} & 67 & 78 & 62 & \f{62} \\

       ANT$^{1\mathrm{x}1}$+SIN & \f{61} & 40 & 39 & 40 & 65 & 76 & 69 & 76 & 67 & 64 & 62 & 55 & 63 & 77 & 59 & 66\\
       ANT$^{1\mathrm{x}1}$ w/o EP & 65 & 46 & 46 & 47 & 66 & 76 & 73 & 75 & 76 & 69 & 70 & 57 & 72 & 79 & 63 & 64 \\[0.5em]
            
       ANT$^{3\mathrm{x}3}$ & 63 & 39 & 40 & \f{39} & 68 & 78 & 73 & 77 & 71 & 66 & 68 & 55 & 69 & 79 & 63 & 64 \\
       ANT$^{3\mathrm{x}3}$+SIN & \f{61} & 43 & 44 & 43 & 71 & \f{74} & 69 & 79 & \f{60} &\f{58} & \f{55} & 56 & 59 & \f{73} & \f{57} & 67 \\
      \end{tabular}
      \caption{Average mean Corruption Error (mCE) obtained by different models on common corruptions from ImageNet-C; lower is better.}
         \label{tab:IN-C3}
  \end{center}
\end{table}

\clearpage
\subsection{MNIST-C results}

\begin{figure*}
    \begin{center}
      \includegraphics[width=0.5\textwidth]{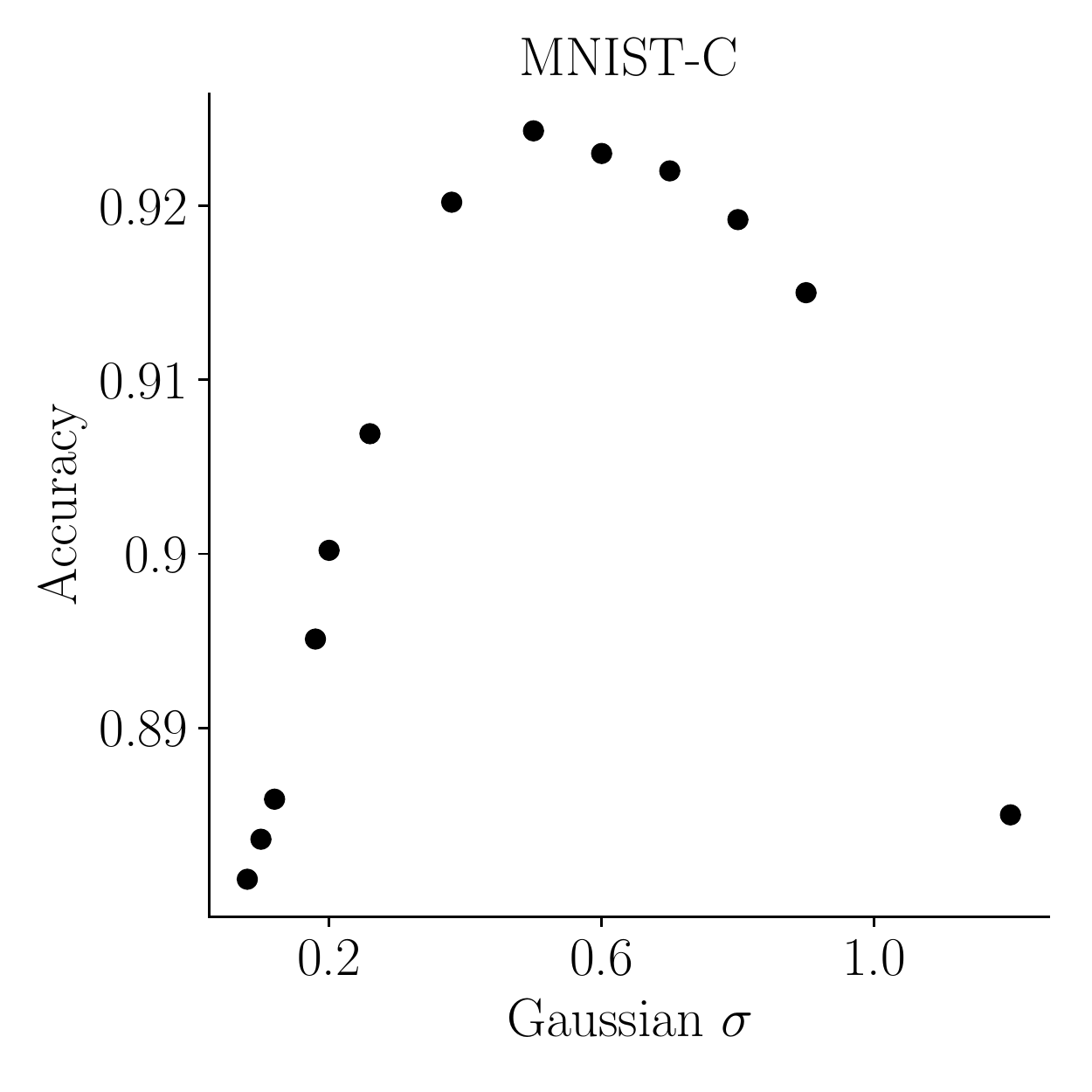}
      \caption{Average accuracy on MNIST-C over all severties and corruptions for different values of sigma $\sigma$ of the Gaussian noise training (GNT) during training. Each point corresponds to one converged training. }
            \label{fig:MNIST_fig_gauss}
      \end{center}
    \end{figure*}
Similar to the ImageNet-C experiments, we are interested how vanilla, adversarially and noise trained models perform on MNIST-C.

The adversarially robust MNIST model by  \citet{DBLP:journals/corr/abs-1805-12514} was trained with a robust loss function and is among the state of the art in certified adversarial robustness. The other baseline models were trained with Adversarial Training in $\ell_2$ (DDN) by \citet{rony2019decoupling} and $\ell_{\infty}$ (PGD) by  \citet{madry2017towards}. Our GNT and ANT$^{1\mathrm{x}1}$ trained versions are trained as described in the main paper and Appendix B.2. The results are shown in Table~\ref{tab:MNIST-C2}. Similar to ImageNet-C, the models trained with GNT and ANT$^{1\mathrm{x}1}$ are significantly better than our vanilla trained baseline. Also, regular adversarial training has severe drops and does not lead to significant robustness improvements.

As for ImageNet and GNT, we have treated $\sigma$ as a hyper-parameter. The accuracy on MNIST-C for different values of $\sigma$ is displayed in Fig.~\ref{fig:MNIST_fig_gauss} and has a maximum around $\sigma=0.5$ like for ImageNet.

\vspace{2cm}

\begin{table*}[ht]\footnotesize
      \begin{center}
      \setlength\tabcolsep{5pt} 
          \begin{tabular}{l c c|c c c c c c c c c c c c c c c} 
           model & \begin{rotate}{45}clean acc\end{rotate} &  \multicolumn{1}{c}{\begin{rotate}{45}mean\end{rotate}} &  \begin{rotate}{45}Shot\end{rotate} &  \begin{rotate}{45}Impulse\end{rotate} &  \begin{rotate}{45}Glass Blur\end{rotate} &  \begin{rotate}{45} Motion Blur\end{rotate} &  \begin{rotate}{45} Shear\end{rotate} &  \begin{rotate}{45} Scale\end{rotate} &  \begin{rotate}{45}Rotate\end{rotate} &  \begin{rotate}{45} Brightness\end{rotate} &  \begin{rotate}{45} Translate\end{rotate} &  \begin{rotate}{45} Stripe \end{rotate} &  \begin{rotate}{45} Fog\end{rotate} &  \begin{rotate}{45} Splatter\end{rotate} &  \begin{rotate}{45} Dotted Line\end{rotate} &  \begin{rotate}{45} Zig Zag\end{rotate} &  \begin{rotate}{45}Canny Edges\end{rotate} \\

             \hline 
             Vanilla & 99.1 & 86.9 & 98 & 96 & 96 & 94 & 98 & 95 & 92 & 88 & 57 & 88 & 50 & 97 & 96 & 86 & 72 \\
             \citep{madry2017towards} & 98.5 & 75.6 & 98 & 55 & 94 & 94 & 97 & 88 & 92 & 27 & 53 & 40 & 63 & 96 & 78 & 74 & 84 \\[0.5em]
             
             Vanilla & 98.8 & 74.3 & 98 & 91 & 96 & 88 & 95 & 80 & 89 & 34 & 45 & 41 & 23 & 96 & 96 & 80 & 63 \\
             \citep{DBLP:journals/corr/abs-1805-12514} & 98.2 & 68.6 & 97 & 65 & 93 & 93 & 94 & 87 & 89 & 11 & 40 & 20 & 25 & 96 & 89 & 61 & 68 \\[0.5em]
             
             Vanilla & 99.5 & 89.8 & 98 & 96 & 95 & 97 & 98 & 96 & 94 & 95 & 61 & 89 & 79 & 98 & 98 & 90 & 63 \\
             DDN Tr \citep{rony2019decoupling} & 99.0 & 87.0 & 99 & 97 & 96 & 94 & 98 & 91 & 93 & 72 & 55 & 92 & 64 & 99 & 98 & 91 & 66 \\[0.5em]
             
             Vanilla & 99.1 & 86.9 & 98 & 96 & 96 & 94 & 98 & 95 & 92 & 88 & 57 & 88 & 50 & 97 & 96 & 86 & 72 \\
             GNT$\sigma_{0.5}$ & 99.3 & 92.4 & 99 & 99 & 98 & 97 & 98 & 95 & 93 & 98 & 56 & 91 & 91 & 99 & 99 & 96 & 78 \\
             ANT$^{1\mathrm{x}1}$ & 99.4 & 92.4 & 99 & 99 & 98 & 97 & 98 & 95 & 93 & 98 & 55 & 89 & 91 & 99 & 99 & 96 & 80 \\

      \end{tabular}
      \caption{Accuracy in percent for the MNIST-C dataset for adversarially robust (\protect{\citep{DBLP:journals/corr/abs-1805-12514}},  \protect{\citep{madry2017towards}}, DDN \protect{\citep{rony2019decoupling}}) and our noise trained models (GNT and ANT$^{1\mathrm{x}1}$). Vanilla always denotes the same network architecture as its adversarially or noise trained counterpart but with standard training. Note that we used the same network architecture as \protect{\citet{madry2017towards}}.}
         \label{tab:MNIST-C2}
  \end{center}
\end{table*}

\clearpage
\subsection{Evaluation of adversarial robustness of models trained via GNT and ANT$^{1\mathrm{x}1}$}
\paragraph{ImageNet}
To evaluate adversarial robustness on ImageNet, we used PGD \citep{madry2017towards} and DDN \citep{rony2019decoupling}. For the $\ell_\infty$ PGD attack, we allowed for 200 iterations with a step size of $0.0001$ and a maximum sphere size of 0.001. For the DDN $\ell_2$ attack, we also allowed for 200 iterations, set the sphere adjustment parameter $\gamma$ to 0.02 and the maximum epsilon to 0.125. We note that for both attacks increasing the number of iterations from 100 to 200 did not make a significant difference in robustness of our tested models. The results on adversarial robustness on ImageNet can be found in the main paper in Table~\ref{tab:adv_robustness_in}.

\paragraph{MNIST}
To evaluate adversarial robustness on MNIST, we also used PGD \citep{madry2017towards} and DDN \citep{rony2019decoupling}. For the $\ell_\infty$ PGD attack, we allowed for 100 iterations with a step size of $0.01$ and a maximum sphere size of 0.1. For the DDN $\ell_2$ attack, we also allowed for 100 iterations, set the sphere adjustment parameter $\gamma$ to 0.05 and the maximum epsilon to 1.5. All models have the same architecture as \citet{madry2017towards}. The results on adversarial robustness on MNIST can be found in Table~\ref{tab:Adv Robustness MNIST}.

\begin{table}[h!]
  \begin{center}
      \begin{tabular}{l c c c} 
           model & clean acc.\,[\%] &  $\ell_2$ acc.\,[\%]  &  $\ell_{\infty}$ acc.\,[\%]  \\

            \hline 
             Vanilla  &   99.1  &   73.2  &   55.8 \\
             GNT$\sigma_{0.5}$  &   99.3  &   89.2  &   73.6 \\
             ANT$^{1\mathrm{x}1}$  &   99.4  &   90.4  &   76.3 \\

      \end{tabular}
      \caption{Adversarial robustness on MNIST on  $\ell_2$  ($\epsilon = 1.5 $) and $\ell_{\infty}$  ($\epsilon = 0.1 $) compared to a Vanilla CNN.}
         \label{tab:Adv Robustness MNIST}
  \end{center}
\end{table}
\clearpage

\subsection{Example images for additive Gaussian noise}
\begin{figure*}
    \begin{center}
      \includegraphics[width=0.8\textwidth]{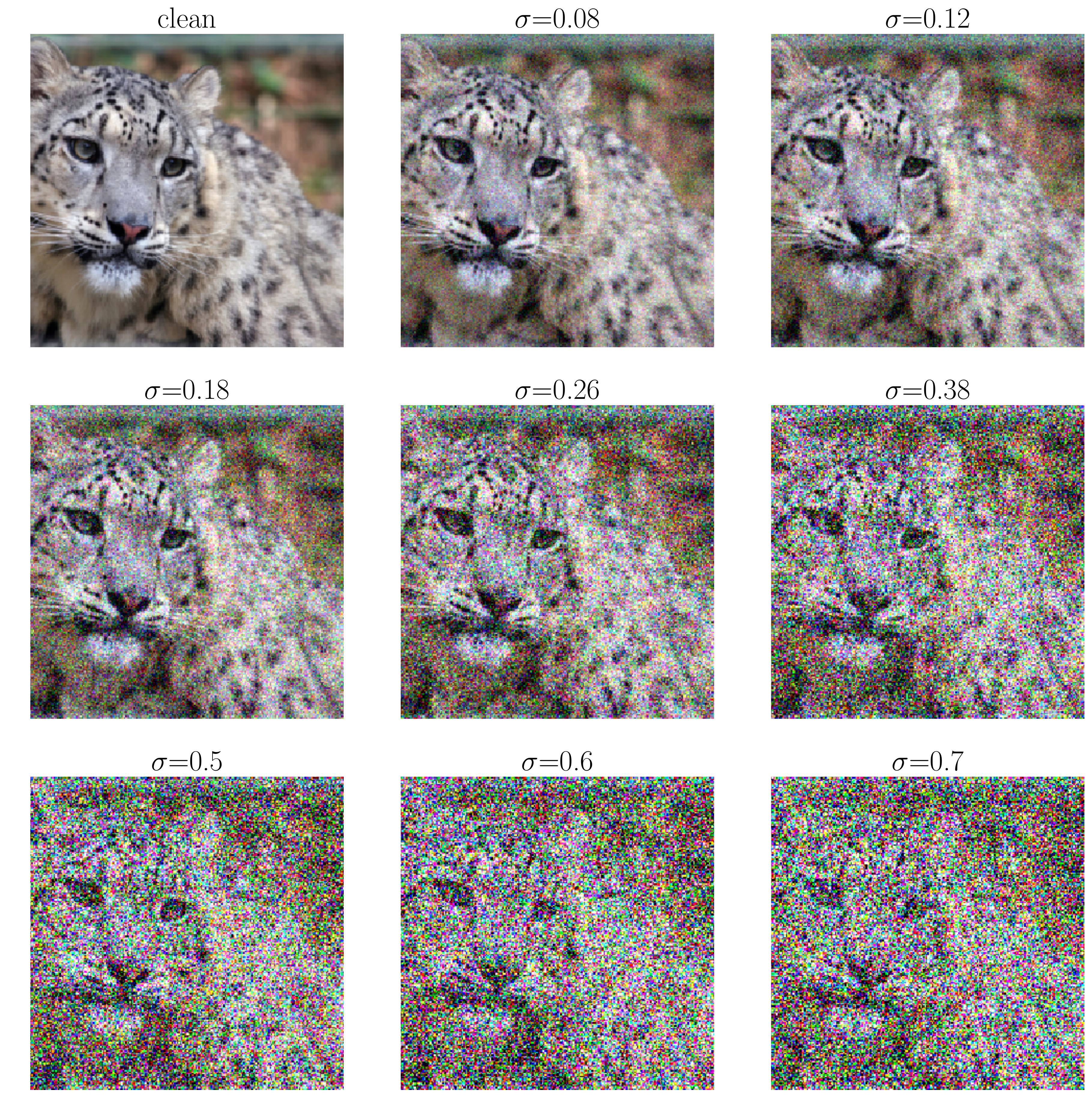}
      \caption{Example images with different $\sigma$-levels of additive Gaussian noise on ImageNet.}
            \label{fig:Example_GN}
      \end{center}
    \end{figure*}

Example images with additive Gaussian noise of varying standard deviation $\sigma$ are displayed in Fig.~\ref{fig:Example_GN}. The considered $\sigma$-levels correspond to those studied in section 4.2.\ in the main paper.

\clearpage 

\subsection{Comparison to Ford et al.}
Ford et al.\ trained an InceptionV3 model from scratch both on clean data from the ImageNet dataset and on data augmented with Gaussian noise \citep{ford2019adversarial}. Since we use a very similar approach, we compare our approach to theirs directly.
The results for comparison on ImageNet both for the vanilla and the Gaussian noise trained model can be found in Table~\ref{tab:ford}. Since we use a pretrained model provided by PyTorch and fine-tune it instead of training a new one, the performance of our vanilla trained model differs from the performance of their vanilla trained model, both on clean data and on ImageNet-C. The accuracy on clean data is displayed in Table~\ref{tab:clean-acc}. 
Another difference between our training and theirs is that we split every batch evenly in clean and data augmented by Gaussian noise with one standard deviation whereas they sample $\sigma$ uniformly between 0 and one specific value.
With our training scheme, we were able to outperform their model significantly on all corruptions except for Elastic, Fog and Brightness. 
\begingroup
\setlength{\tabcolsep}{3pt} 
\begin{table}[ht]\footnotesize
  \begin{center}
      \begin{tabular}{l|c|c c c|c c c c} 
        &  &  \multicolumn{3}{c|}{Noise (Compressed)}  &  \multicolumn{4}{c}{Blur (Compressed)} \\
        model              & All & Gaussian & Shot& Impulse& Defocus& Glass   & Motion &  Zoom \\
        \hline 
        Vanilla InceptionV3 \citep{ford2019adversarial} & 38.8 & 36.6     & 34.3     & 34.7     & 31.1    & 19.3    & 35.3     & 30.1 \\
        Gaussian ($\sigma=0.4$) \citep{ford2019adversarial} & 42.7 & 40.3 & 38.8 & 37.7 & 32.9 & 29.8 & 35.3 & 33.1 \\[0.5em]
        
        Vanilla InceptionV3 [ours] & 41.6 & 42.0 & 40.3 & 38.5 & 33.5 & 27.1 & 36.1 & 28.8 \\
        GNT$\sigma_{0.4}$ [ours] & 49.5 & 60.8 & 59.6 & 59.4 & 43.8 & 37.0 & 42.8 & 38.4\\
        GNT$\sigma_{0.5}$ [ours] & \f{50.2} & \f{61.6} & \f{60.9} & \f{60.8} & \f{44.6} & \f{37.3} & \f{44.0} & \f{39.3}  \\

      \end{tabular}
\newline
\vspace*{0.5 cm}
\newline
      \begin{tabular}{l|c c c c|c c c c} 
        &   \multicolumn{4}{c|}{Weather (Compressed)}  &  \multicolumn{4}{c}{Digital (Compressed)} \\
        model               & Snow   & Frost & Fog      &Brightness & Contrast & Elastic & Pixelate &  JPEG \\
        \hline 
        Vanilla InceptionV3 \citep{ford2019adversarial}   & 33.1     & 34.0     & 52.4     & 66.0     & 35.9     & 47.8    & 38.2     & 50.0  \\
        Gaussian ($\sigma=0.4$) \citep{ford2019adversarial} & 36.6 & 43.5& \f{52.3} & \f{67.1} & 35.8 & \f{52.2}& 47.0 & 55.5 \\[0.5em]
        
        Vanilla InceptionV3 [ours] & 33.5 & 39.6 & 42.2 & 64.2 & 41.0 & 43.5 & 57.4 & 56.9 \\
        GNT$\sigma_{0.4}$ [ours] & 35.6 & 43.7 & 43.3 & 64.8 & 43.0 & 49.0 & 59.3 & 61.7\\
        GNT$\sigma_{0.5}$ [ours] & \f{37.1} & \f{44.2} & 43.6 & 64.6 & \f{43.3} & 49.4 & \f{59.6} & \f{61.9} \\
      \end{tabular}
      \captionsetup{justification=centering}
      \caption{ImageNet-C accuracy for InceptionV3.}
         \label{tab:ford}
  \end{center}
\end{table}
\endgroup

\begin{table*}[ht]
      \begin{center}
          \begin{tabular}{l  c } 
           model & clean accuracy [\%]  \\

            \hline 
             Vanilla InceptionV3 \citep{ford2019adversarial} &  75.9 \\
             Gaussian ($\sigma=0.4$) \citep{ford2019adversarial} & 74.2 \\[0.5em]
             
             Vanilla InceptionV3 [ours]  & 77.2 \\
             GNT$\sigma_{0.4}$ [ours] & 78.1 \\
             GNT$\sigma_{0.5}$ [ours] & 77.9 \\

      \end{tabular}
      \captionsetup{justification=centering}
      \caption{Accuracy on clean data for differently trained models.}
         \label{tab:clean-acc}
  \end{center}
\end{table*}

\clearpage 

\subsection{Visualization of images with different perturbation sizes}
In the main paper, we measure model robustness by calculating the median perturbation size $\epsilon^*$ and report the results in Table~\ref{tab:jt}. To provide a better intuition for the noise level in an image for a particular $\epsilon^*$, we display example images in Fig.~\ref{Figure:ex_tab_3}.

\begingroup
\begin{figure}[h!]
\setlength\tabcolsep{0.0pt} 
\renewcommand{\arraystretch}{0.1}
\centering
\begin{tabular}{c c c c}
 & Gaussian Noise & Uniform Noise & Adversarial Noise \\[-1.5ex]
\begin{turn}{90}$\;\;\;\;\;\;\;\;\;\;$Vanilla RN50\end{turn} & \subfloat{\includegraphics[width=0.27\textwidth]{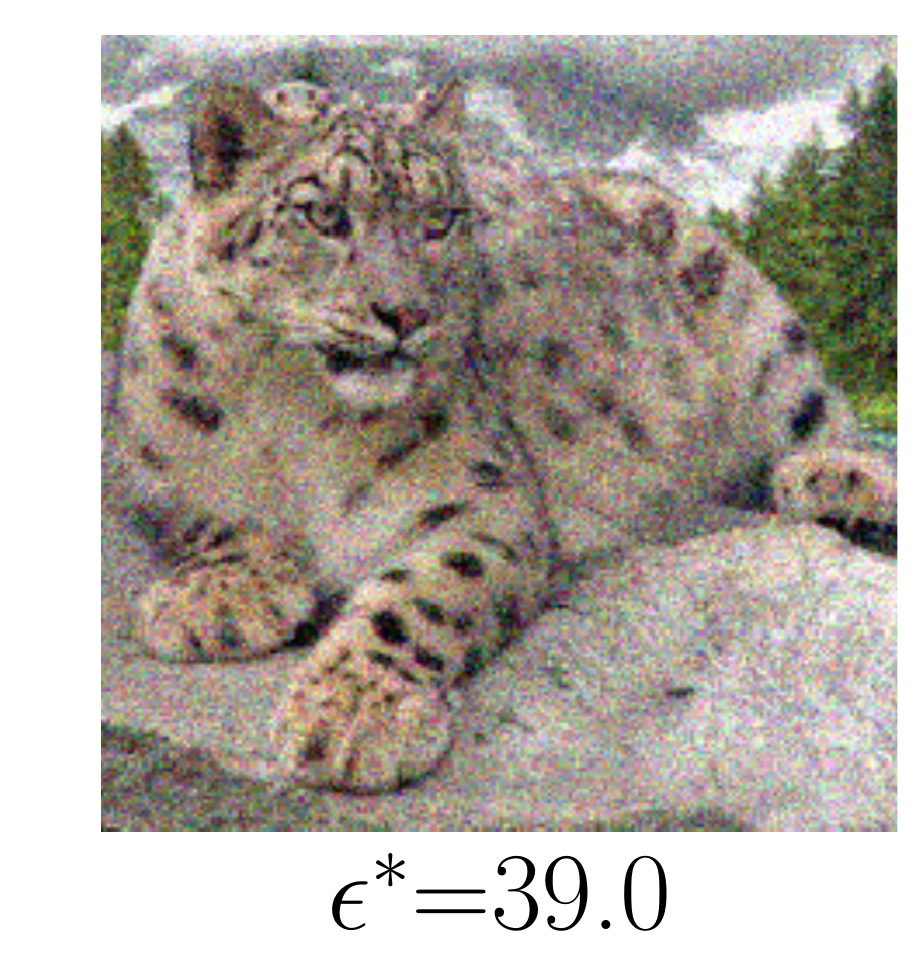}}
& \subfloat{\includegraphics[width=0.27\textwidth]{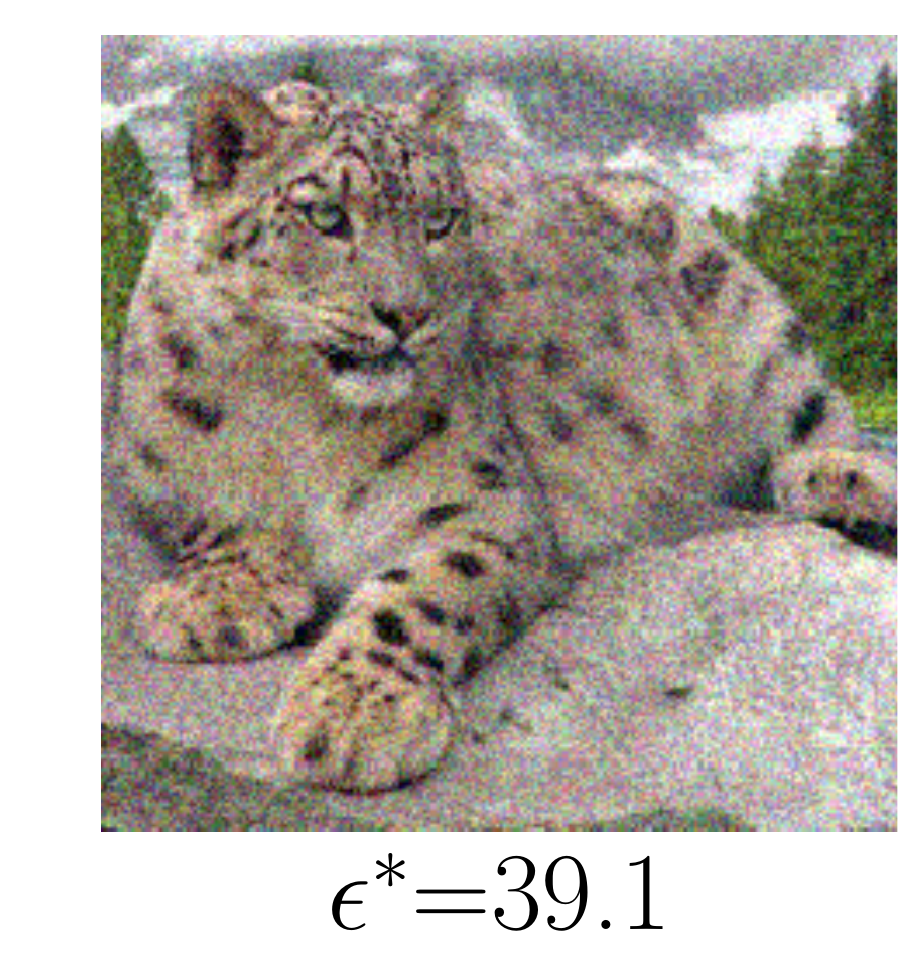}}
& \subfloat{\includegraphics[width=0.27\textwidth]{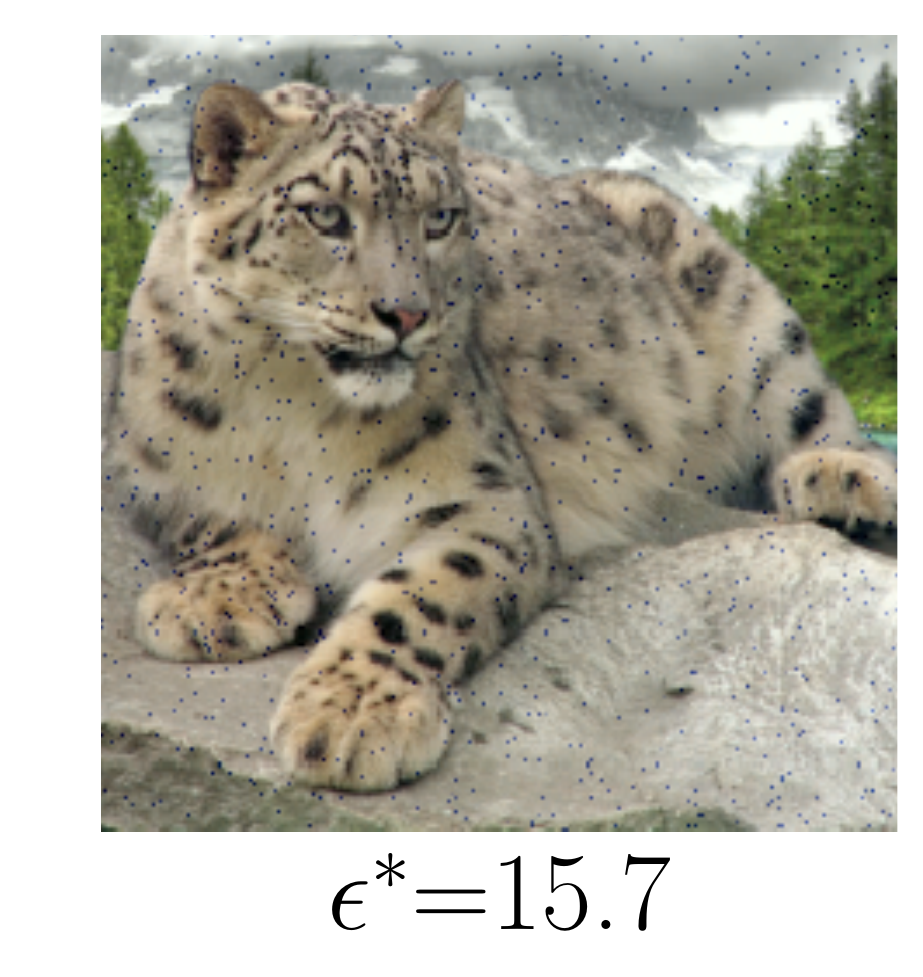}} \\[-2.ex]

\begin{turn}{90}$\;\;\;\;\;\;\;\;\;\;\;\;\;\;$GNT$\sigma_{0.5}$\end{turn} & \subfloat{\includegraphics[width=0.27\textwidth]{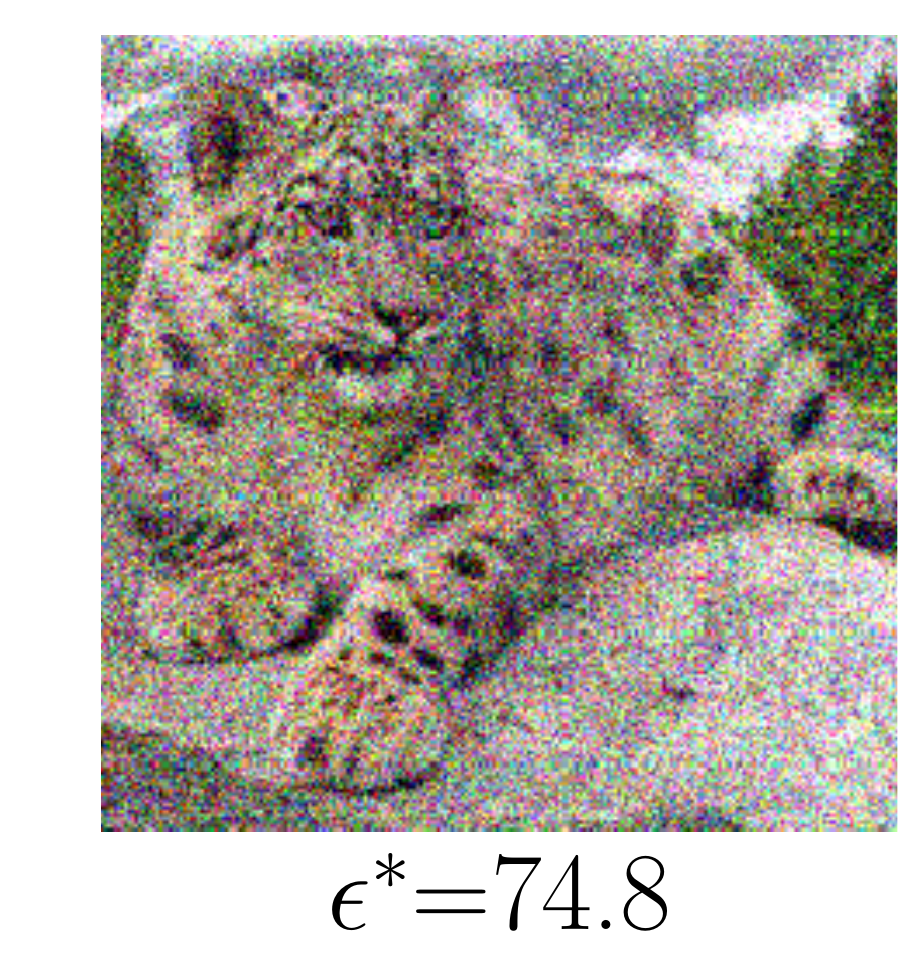}}
& \subfloat{\includegraphics[width=0.27\textwidth]{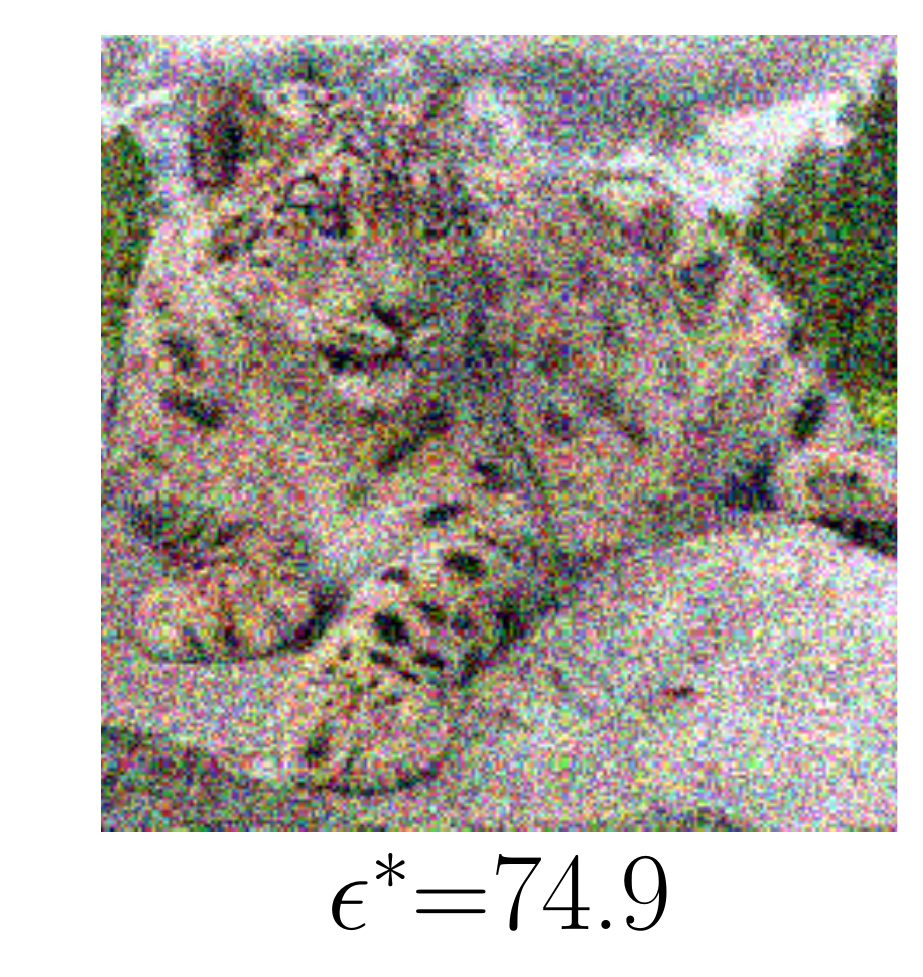}}
& \subfloat{\includegraphics[width=0.27\textwidth]{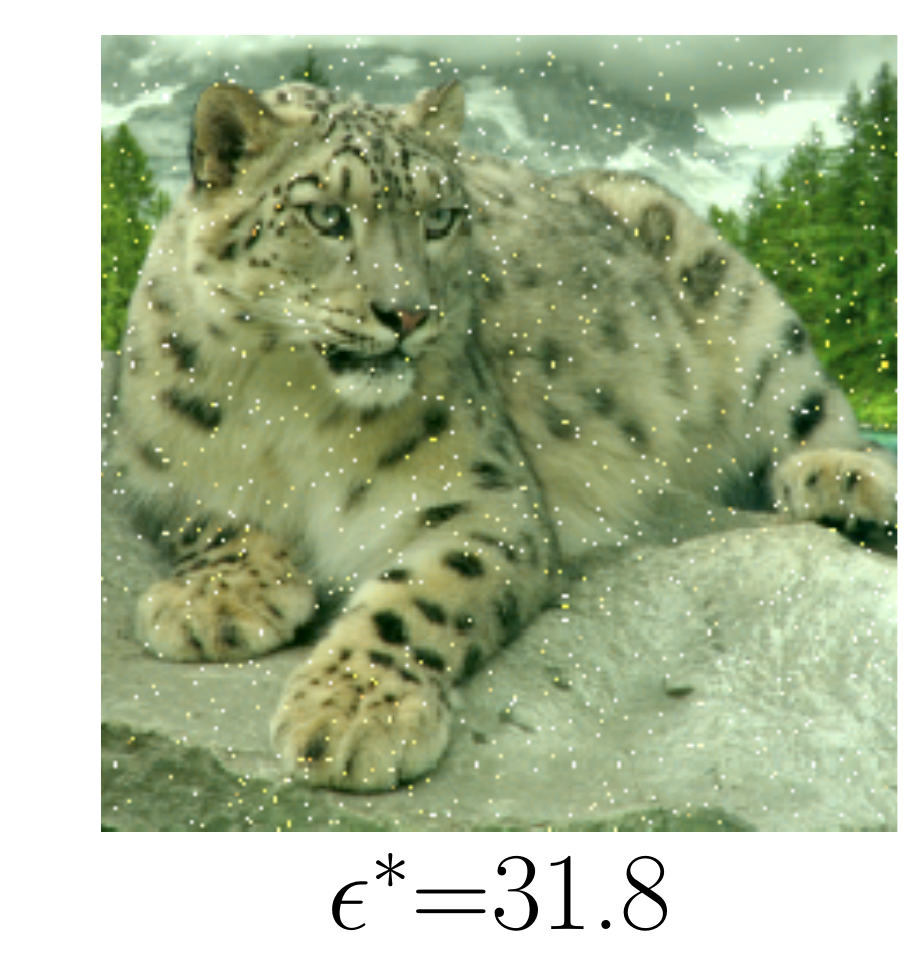}} \\[-2.ex]

\begin{turn}{90}$\;\;\;\;\;\;\;\;\;\;\;\;\;\;$GNT$_{\mathrm{mult}}$\end{turn} & \subfloat{\includegraphics[width=0.27\textwidth]{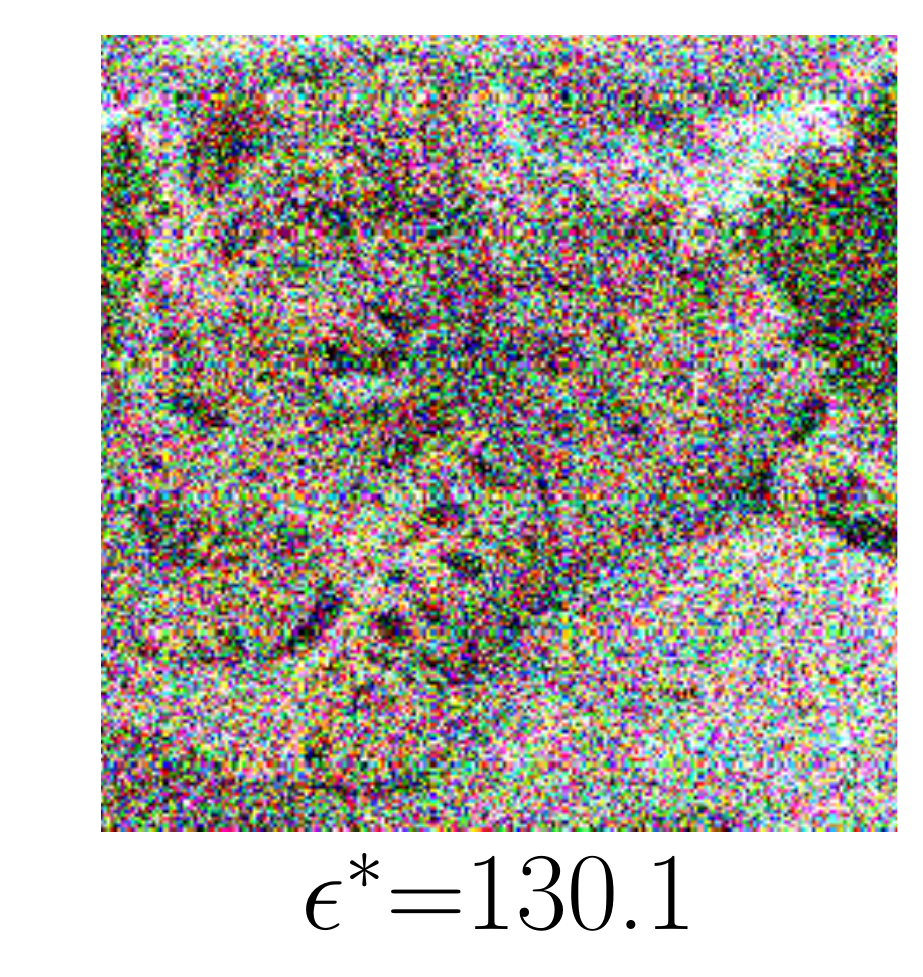}}
& \subfloat{\includegraphics[width=0.27\textwidth]{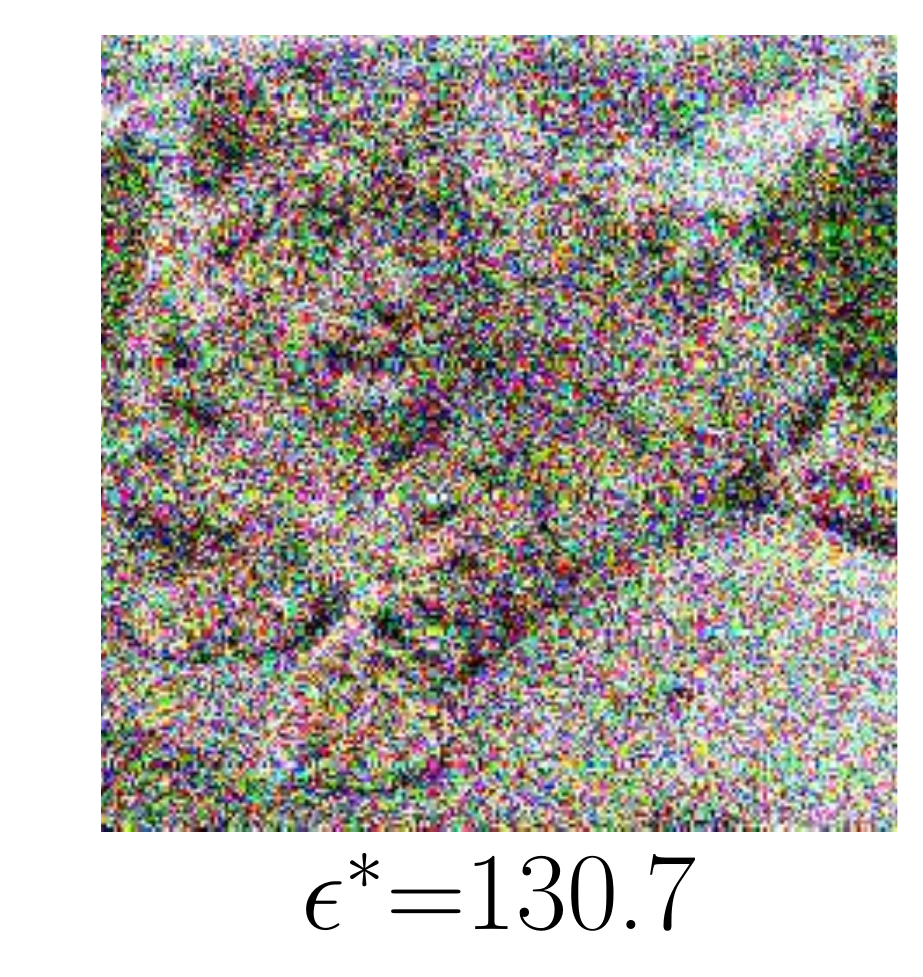}}
& \subfloat{\includegraphics[width=0.27\textwidth]{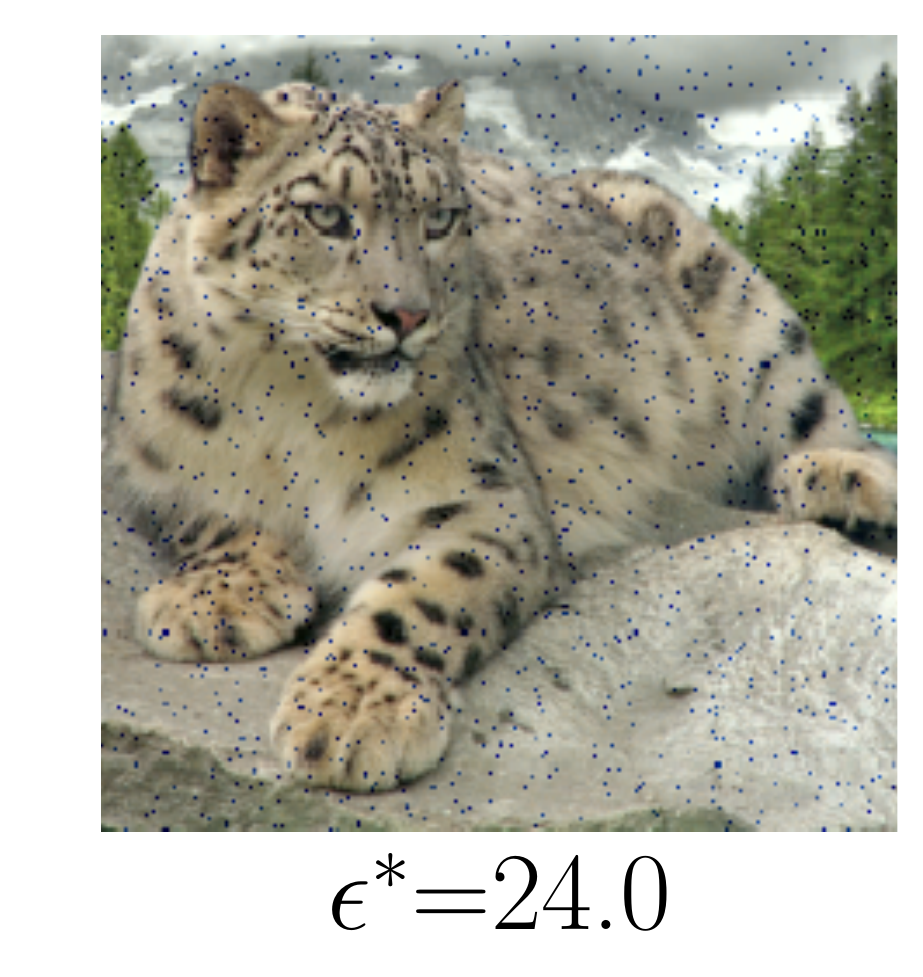}} \\[-2.ex]

\begin{turn}{90}$\;\;\;\;\;\;\;\;\;\;\;\;\;\;\;\;\;\;$ANT$^{1\mathrm{x}1}$\end{turn} 
& \subfloat{\includegraphics[width=0.27\textwidth]{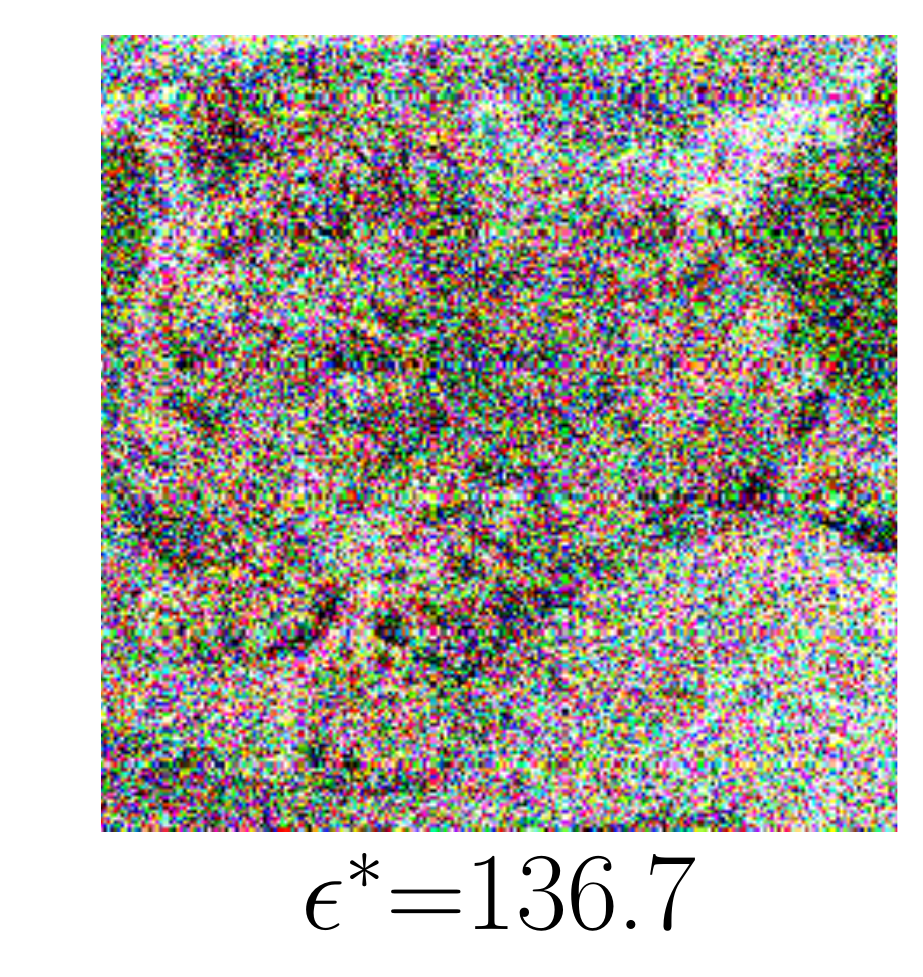}}
& \subfloat{\includegraphics[width=0.27\textwidth]{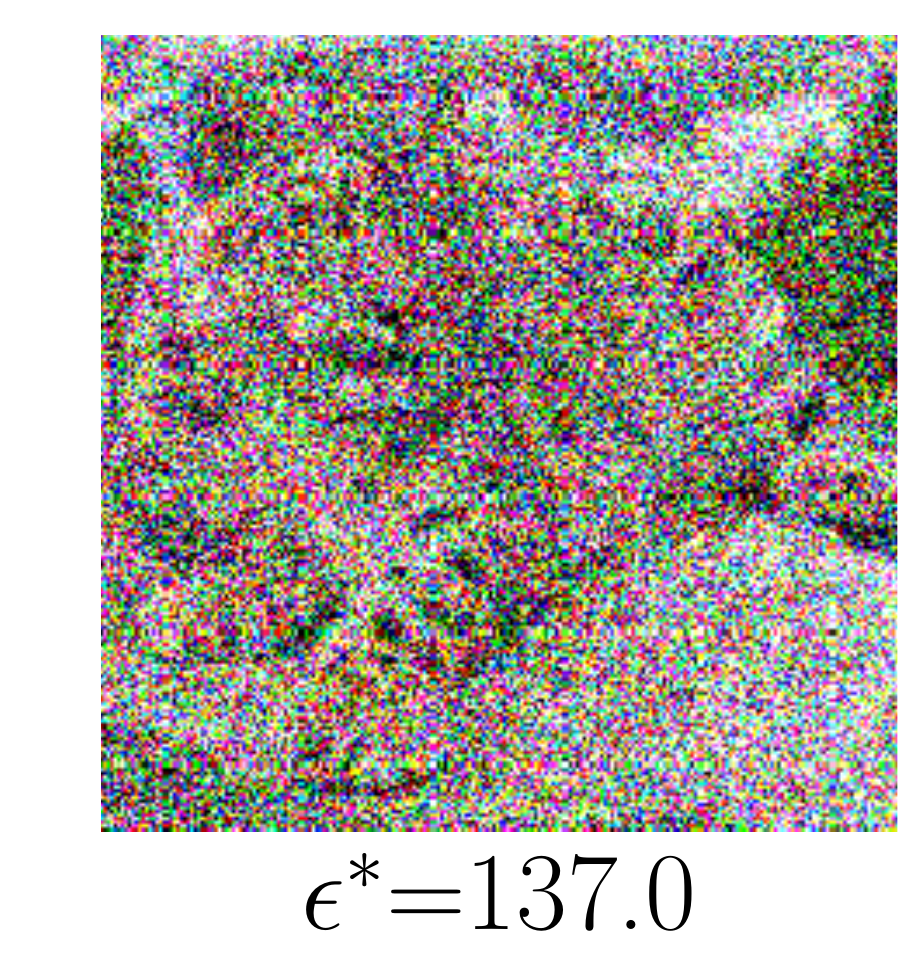}}
& \subfloat{\includegraphics[width=0.27\textwidth]{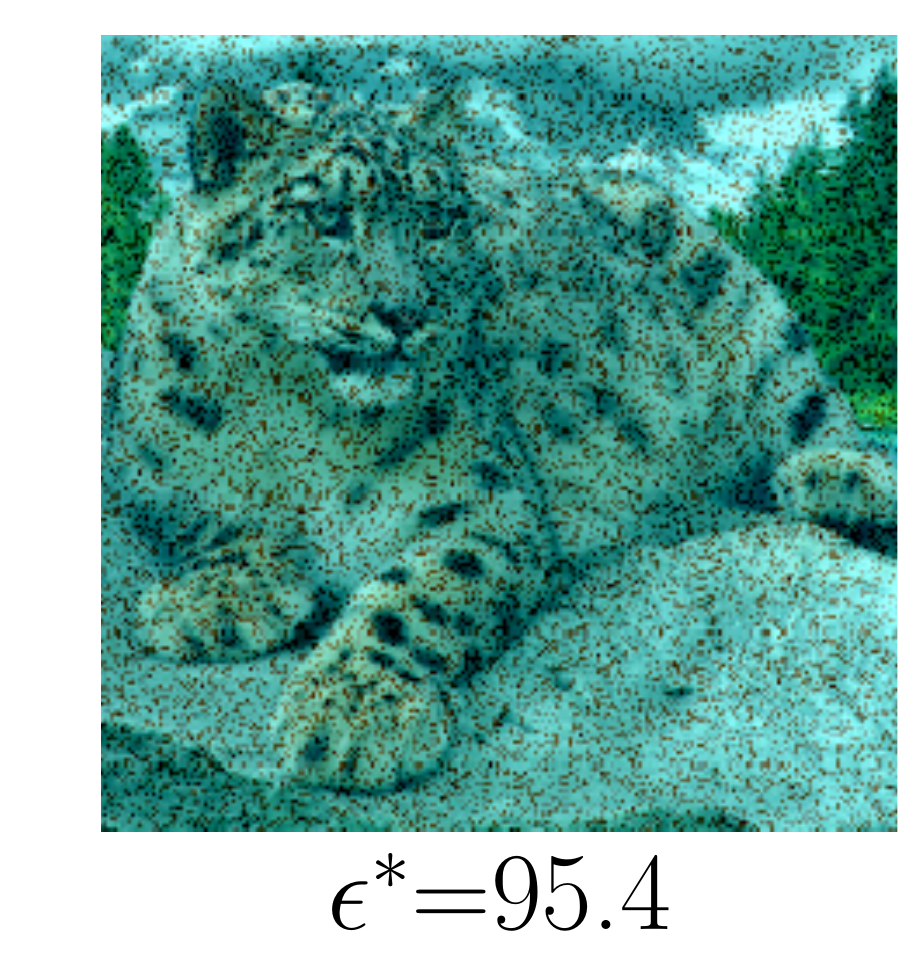}} \\
\end{tabular}
 \caption{Example images for the different perturbation sizes $\epsilon^*$ and different noise types on ImageNet corresponding to the $\epsilon^*$ values in Table~\ref{tab:jt} in the main paper.}
 \label{Figure:ex_tab_3}
\end{figure}
\endgroup

\clearpage 

\subsection{Visualization of Posterize vs JPEG}

\begin{figure*}
    \begin{center}
      \includegraphics[width=0.8\textwidth]{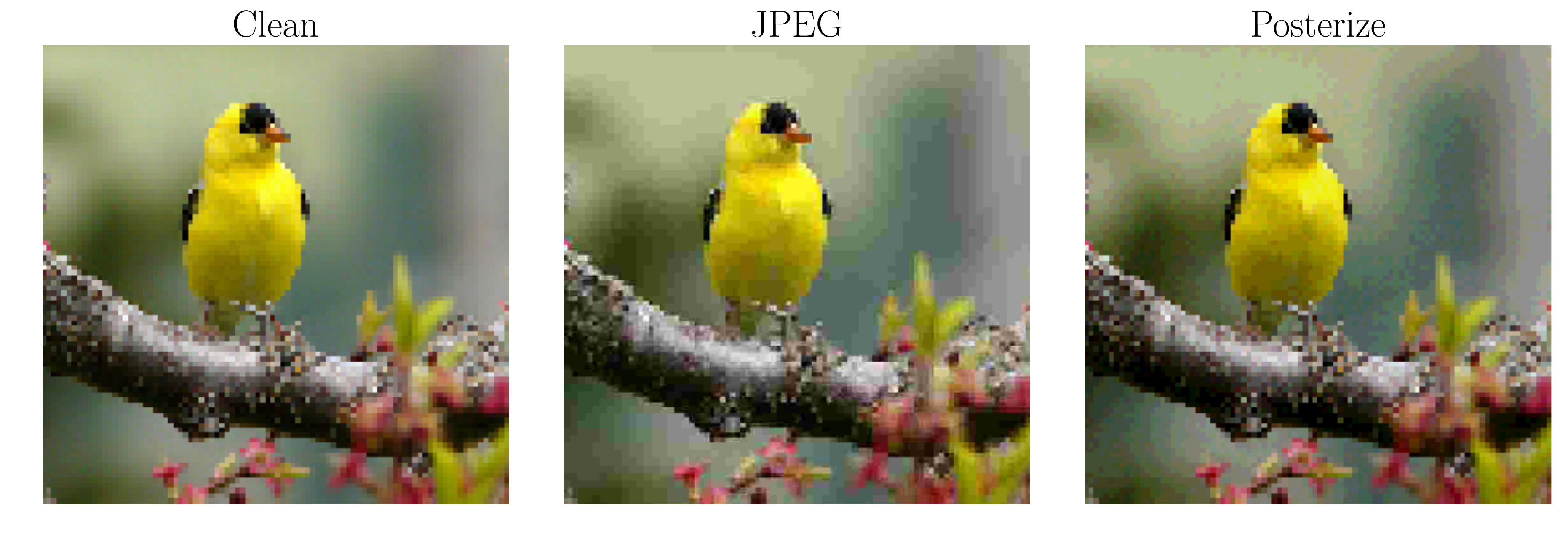}
      \caption{Example images for the JPEG compression from ImageNet-C and the \texttt{PIL.ImageOps.Posterize} operation.}
                \label{fig:posterize}
      \end{center}
    \end{figure*}

AugMix \citep{hendrycks2020augmix} uses Posterize as one of their operations for data augmentation during training. In Fig.~\ref{fig:posterize}, we show the visual similarity between the Posterize operation and the JPEG corruption from ImageNet-C.

\subsection{Additional results}

\subsubsection{Adversarial Noise Training with a DenseNet121 architecture}
To test in how far our results generalize to other backbones, we have trained a DenseNet121 model with ANT$^{1\mathrm{x}1}$. The DenseNet121 model was finetuned from the checkpoint provided by torchvision. 
A DenseNet121 has $7.97886\cdot10^6$ trainable parameters whereas a ResNet50 has $2.5557\cdot10^7$.
Our results and a comparison to ANT$^{1\mathrm{x}1}$ with a ResNet50 model is shown in Table~\ref{densenet}: ANT$^{1\mathrm{x}1}$ increases robustness on full ImageNet-C and ImageNet-C without noises for the DenseNet121 model, showing that adversarial noise training generalizes to other backbones.
    
\begin{table}
  \begin{center}
  \setlength\tabcolsep{0.5pt} 
      \begin{tabular}{l c c c c c} 
         & IN & \multicolumn{2}{c}{IN-C} & \multicolumn{2}{c}{IN-C w/o noises} \\
        model    & clean acc. & $\;$ Top-1$\;$ & $\;$Top-5$\;$ &$\;$ Top-1$\;$ &$\;$ Top-5\\
        \hline 
        Vanilla RN50 & 76.1 & 39.2 & 59.3  & 42.3 & 63.2\\
       ANT$^{1\mathrm{x}1}$ RN50  & 76.0 & (\g{51.1})  & (\g{72.2})   &   47.7   & 68.8 \\[0.5em]
       Vanilla DN121 & 74.4 & 42.1 & 63.4 & 44.0 & 65.5  \\
       ANT$^{1\mathrm{x}1}$ DN121  & 74.3 & 50.3 & 71.6 & 46.8 & 68.3 \\
      \end{tabular}
      \caption{Average accuracy on clean data, average Top-1 and Top-5 accuracies on full ImageNet-C and ImageNet-C without the noise category (higher is better); all values in percent. We compare the results obtained by ANT$^{1\mathrm{x}1}$ for a ResNet50 (RN50) architecture to a DenseNet121 (DN121) architecture.}
         \label{densenet}
  \end{center}
\end{table}

\subsubsection{Results for different parameter counts of the noise generator}
Here, we study the effect of different parameter counts of the adversarial noise generator on ANT$^{1\mathrm{x}1}$. We provide the results in Table~\ref{tab:ablation_parameters}. We indicate the depth of the noise generator with a subscript. All experiments in this paper apart from this ablation study were performed with a default depth of 4 layers. We observe that while depth is a tunable hyper-parameter, the performances of ANT$^{1\mathrm{x}1}$ with the studied noise generators do not differ by a lot. Only the most shallow noise generator with a depth of one layer and only 12 trainable parameters results in a roughly 1\% lower accuracy than its deeper counterparts. We note that a GNT$\sigma_{0.5}$ model has an accuracy of 49.4\% on full ImageNet-C and an accuracy of 47.1\% on ImageNet-C without noises which roughly corresponds to the respective accuracies of ANT$^{1\mathrm{x}1}$ with the most shallow noise generator.

\begin{table}
  \begin{center}
  \setlength\tabcolsep{0.5pt} 
      \begin{tabular}{l c c c c} 
        &  Number of & IN & IN-C & IN-C w/o noises \\
        model    & $\;\;$parameters$\;\;$ & clean acc. & $\;$ Top-1$\;$ & $\;$ Top-1$\;$\\[0.5em]
        \hline 
        Vanilla RN50 &  - & 76.1 & 39.2 & 42.3\\
       ANT$^{1\mathrm{x}1}$ RN50 NG$_1$ & 12 & 75.1 & (\g{49.5}) & 46.6\\
       ANT$^{1\mathrm{x}1}$ RN50 NG$_2$ & 143 &  75.5 & (\g{50.8}) & 47.2 \\
       ANT$^{1\mathrm{x}1}$ RN50 NG$_3$ & 563 & 75.3 & (\g{50.7}) & 47.2 \\
       ANT$^{1\mathrm{x}1}$ RN50 NG$_4$ & 983 & 76.0 & (\g{51.1})  &  47.7 \\
       ANT$^{1\mathrm{x}1}$ RN50 NG$_5$ & 1403 & 74.0 &  (\g{50.7}) & 47.0\\
      \end{tabular}
      \caption{Number of trainable parameters of different noise generators, average accuracy on clean data, ImageNet-C and ImageNet-C without the noise category (higher is better); all values in percent. We compare the results obtained by ANT$^{1\mathrm{x}1}$ with noise generators of different depth. Note that a depth of 4 layers was used in all experiments in this paper apart from this ablation study.}
         \label{tab:ablation_parameters}
  \end{center}
\end{table}

\end{document}